\def\year{2021}\relax
\documentclass[letterpaper]{article} % DO NOT CHANGE THIS

\usepackage{aaai21}  % DO NOT CHANGE THIS
\usepackage{times}  % DO NOT CHANGE THIS
\usepackage{helvet} % DO NOT CHANGE THIS
\usepackage{courier}  % DO NOT CHANGE THIS
\usepackage[hyphens]{url}  % DO NOT CHANGE THIS
\usepackage{graphicx} % DO NOT CHANGE THIS
\usepackage{natbib}  % DO NOT CHANGE THIS AND DO NOT ADD ANY OPTIONS TO IT
\usepackage{caption} % DO NOT CHANGE THIS AND DO NOT ADD ANY OPTIONS TO IT
\usepackage[utf8]{inputenc} % allow utf-8 input
\usepackage[T1]{fontenc}    % use 8-bit T1 fonts
\usepackage{hyperref}       % hyperlinks
\usepackage{url}            % simple URL typesetting
\usepackage{booktabs}       % professional-quality tables
\usepackage{amsfonts}       % blackboard math symbols
\usepackage{nicefrac}       % compact symbols for 1/2, etc.
\usepackage{microtype}      % microtypography

\usepackage[switch]{lineno}
\usepackage{amsmath}
\usepackage{amsthm}
\usepackage{psfrag}
\usepackage{graphicx}
\usepackage{amssymb}
\usepackage{subfigure}
\usepackage{algorithmic}
\usepackage{algorithm}
\usepackage{cases}
\usepackage{ulem}
\usepackage{amsmath,amssymb}
\usepackage{psfrag}
\usepackage{epsfig}
\usepackage{graphics}
\usepackage{color,url}
\usepackage{multirow}
\usepackage{threeparttable}%%%LBL
\usepackage{booktabs}%%%LBL
\usepackage{slashbox}

\usepackage{textcomp}
\usepackage{epsf}
\usepackage{epstopdf}
\usepackage{psfrag}
\usepackage{epsfig}
\usepackage{multirow}
\usepackage{threeparttable}
\usepackage{booktabs}
\usepackage{bm}
\usepackage{float}
\usepackage{appendix}
\usepackage[Symbol]{upgreek}
\usepackage{caption,setspace}
\usepackage{amsmath}
\usepackage{psfrag}
\newtheorem{theorem}{Theorem}

\newtheorem{remark}{Remark}
\newtheorem{definition}{Definition}

\title{Deforming the Loss Surface to Affect the Behaviour of the Optimizer}
\author {
     Liangming Chen,\textsuperscript{\rm 1}
     Long Jin,\textsuperscript{\rm 1} \thanks{Corresponding author}
     Xiujuan Du, \textsuperscript{\rm 2}
     Shuai Li,\textsuperscript{\rm 1}
     Mei Liu \textsuperscript{\rm 1}\\
}
\affiliations {
    \textsuperscript{\rm 1} School of Information Science and Engineering, Lanzhou University\\
    \textsuperscript{\rm 2} School of Computer Science and Technology, Qinghai Normal University\\
    lmchen@foxmail.com, jinlongsysu@foxmail.com, 1070172063@qq.com, lishuai@lzu.edu.cn, mliu@lzu.edu.cn
}
\begin{document}
\maketitle
%\linenumbers
\begin{abstract}
In deep learning, it is usually assumed that the optimization process is conducted on a shape-fixed loss surface. Differently, we first propose a novel concept of deformation mapping in this paper to affect the behaviour of the optimizer. Vertical deformation mapping (VDM), as a type of deformation mapping, can make the optimizer enter a flat region, which often implies better generalization performance. Moreover,  we design various VDMs, and further provide their contributions to the loss surface. After defining the local M region, theoretical analyses show that deforming the loss surface can enhance the gradient descent optimizer's ability to filter out sharp minima. With visualizations of loss landscapes,  we evaluate the flatnesses of minima obtained by both the original optimizer and optimizers enhanced by VDMs on CIFAR-$100$. The experimental results show that VDMs do find flatter regions. Moreover, we compare popular convolutional neural networks enhanced by VDMs with the corresponding original ones on ImageNet, CIFAR-$10$, and CIFAR-$100$. The results are surprising: there are significant improvements on all of the involved models equipped with VDMs. For example, the top-$1$ test accuracy of ResNet-$20$ on CIFAR-$100$ increases by $1.46\%$, with insignificant additional computational overhead.
\end{abstract}

\section{Introduction}\label{intro}
\begin{figure}[t]
	\centering
	\subfigure[Original, i.e. $\check{\ell} = \ell, \check{\boldsymbol{p}} = \boldsymbol{p}$ (filled contour view)]{
		\label{filter_sim_a}
		\includegraphics[width=1.350123in]{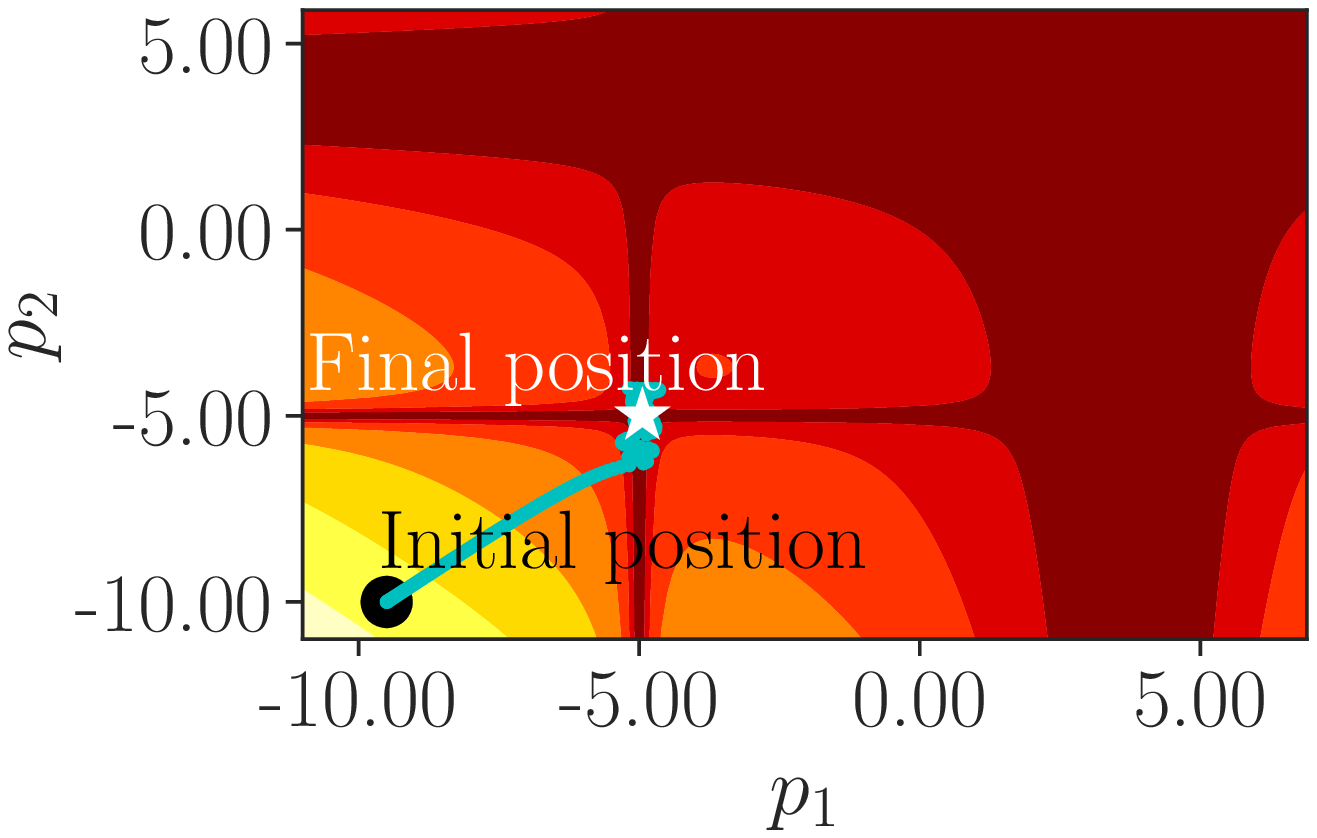}
	}
	\subfigure[Original ($3$D view)]{
	\label{filter_sim_c}
	\includegraphics[width=1.57123in]{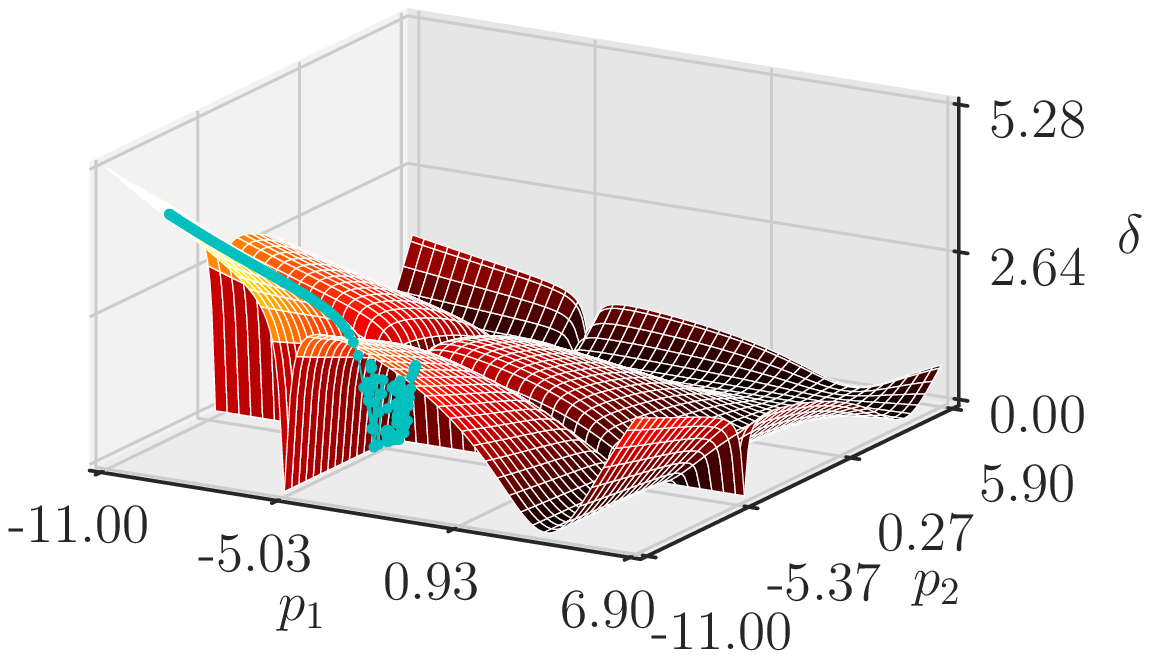}
}
\quad
	\subfigure[Deformed by AP(10, 1, 1, 1) (filled contour view)]{
		\label{filter_sim_b}
		\includegraphics[width=1.350123in]{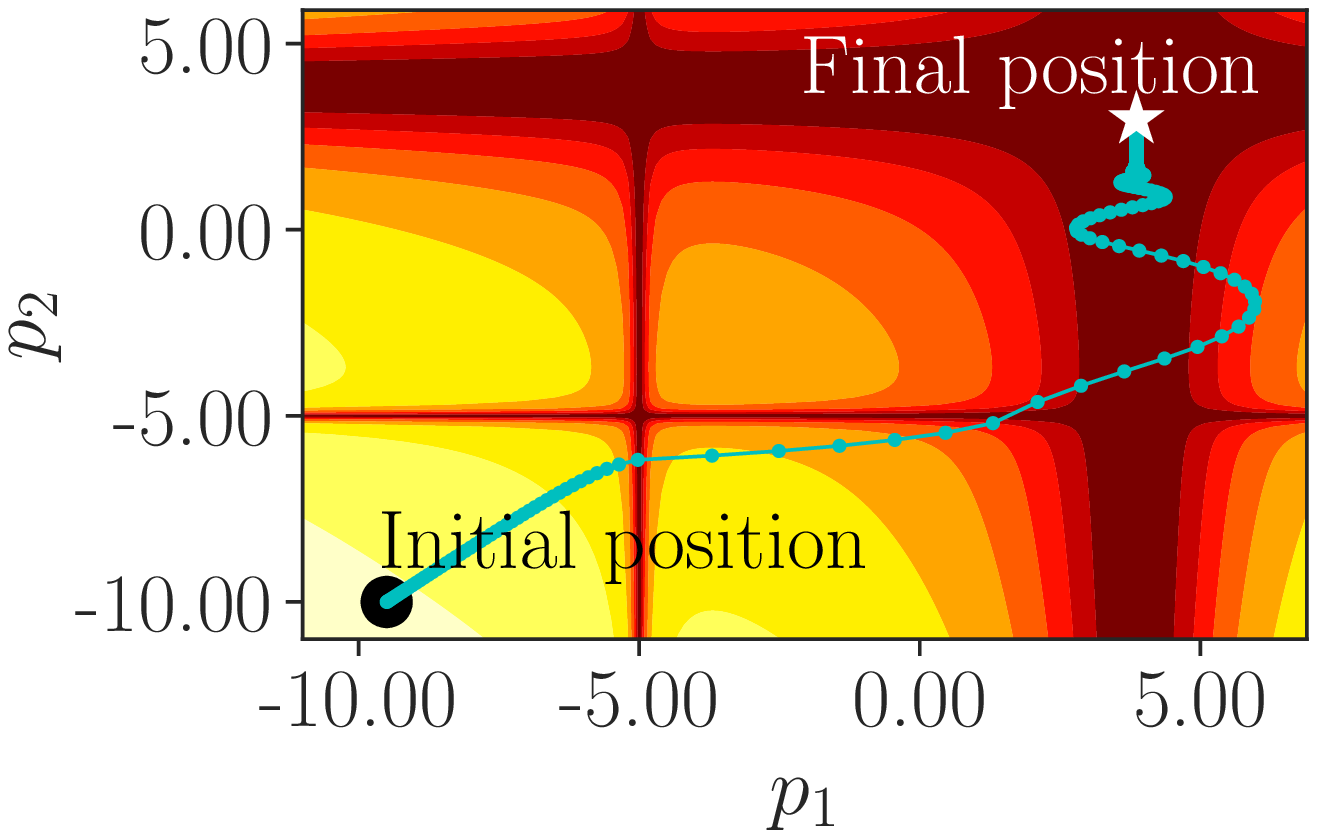}
	}
	\subfigure[Deformed ($3$D view)]{
		\label{filter_sim_d}
		\includegraphics[width=1.570123in]{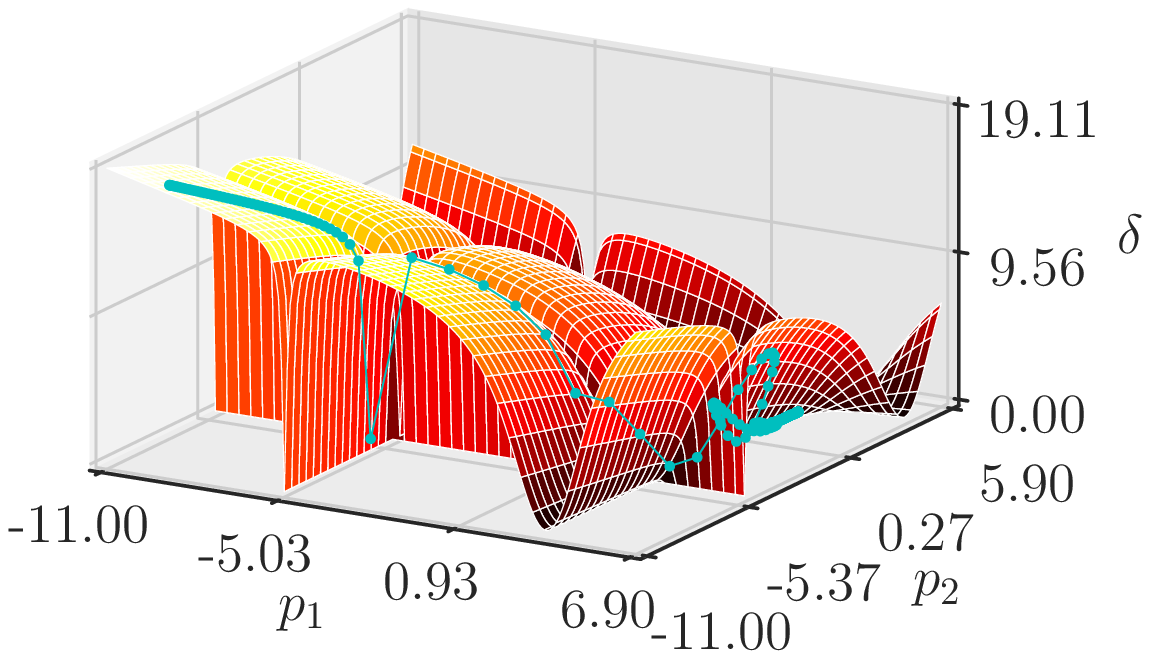}
	}
	\caption{Deformation for a simulated loss surface with two parameters to illustrate that the VDM has the capability to filter out the sharp minimum. Note that the region with high loss has a small degree of deformation, and the region with low loss has a large degree of deformation. }
	\label{filter_sim}
\end{figure}
%\begin{figure}[htbp]
%    \centering
%    \subfigure[Original (filled contour view)]{
%    \label{filter_sim_a}
%        \includegraphics[width=2.5in]{filter_sim-1.eps}
%    }
%    \subfigure[Deformed (filled contour view)]{
%    \label{filter_sim_b}
%	\includegraphics[width=2.5in]{filter_sim-2.eps}
%    }
%    \quad    %? \quad ???
%    \subfigure[Original ($3$D view)]{
%    \label{filter_sim_c}
%    	\includegraphics[width=2.5in]{filter_sim-3.eps}
%    }
%    \subfigure[Deformed ($3$D view)]{
%    \label{filter_sim_d}
%	\includegraphics[width=2.5in]{filter_sim-4.eps}
%    }
%    \caption{Deformation on a simulated loss surface to illustrate that the VDM has the capability to filter out the sharp minimum.}
%\label{filter_sim}
%\end{figure}

%The optimization problem in deep learning is considered to be a non-convex problem with high dimension. 
The training processes of deep learning models can be deemed as searching minima on a high dimensional loss surfaces \cite{fort2019goldilocks}. However, the original shape of the loss surface may not always be conducive to finding well-behaved parameters, for example, parameters with good generalization performance. Although various optimization methods may obtain low training losses, the testing results are diverse \cite{shin2020hlhlp}. A promising direction is to link the generalization ability with the flatness near the critical point \cite{dziugaite2020revisiting, wang2018identifying, wu2017towards, sagun2018empirical, keskar2019large}.
%Existing optimization methods often focus on finding the minimum of the cost function through parameter updating methods.
Some research indicates that although the sharpness alone is not enough to predict the generalization performance accurately, there is still a trend that flat minima generalize better than that of sharp ones \cite{keskar2019large, he2019asymmetric, dinh2017sharp, NIPS2017_7176}. We propose deformation mapping to deform the loss surface to make the optimizer fall into flat minima more easily, supported by both experiments and theoretical analyses.

For deep learning models, there are numerous parameters to enable the loss value to decrease along the negative gradient direction. Even if some of the minima are escaped, the optimizer still has the capability to obtain tiny enough loss value \cite{pennington2017geometry}. To provide an intuitive understanding, numerical simulations with two parameters are performed in Fig. \ref{filter_sim}. Besides, to evaluate the performance of the proposed deforming function, in practice, comparative classification experiments are conducted on ImageNet \cite{russakovsky2015imagenet}, CIFAR-$10$ \cite{Krizhevsky09learningmultiple}, and CIFAR-$100$ \cite{Krizhevsky09learningmultiple} for popular convolutional neural networks (CNNs). Experimental results show that the proposed VDMs improve the test performances of all the participating models significantly. It is worth mentioning that the proposed VDMs bring little additional calculation and memory burden. Our contributions in this paper are summarized as follows:
\begin{itemize}
	\item  The concept of deforming the loss surface to affect the behaviour of the optimizer is proposed for the first time, to the best of our knowledge. In order to achieve the purpose of deforming loss surface, deformation mapping is proposed, along with conditions it should meet.
	
	\item  As one kind of deformation mapping, vertical deformation mapping (VDM) is proposed, along with its deformation characteristics on the loss surface. Specific examples of the VDMs, i.e., arctan-power-based (AP) VDM and log-exp-based (LE) VDM, are given, as well as their contributions to loss surfaces.
	
	\item  We provide theoretical analyses that the vertical-deformation-mapping-equipped GD can find flat minima. One-dimensional case and high-dimensional case are analyzed theoretically. Specifically, the landscape near the solution obtained by the vertical-deformation-mapping-equipped optimizer is flatter, because some sharp minima are filtered out during the training process. Besides, visualizing experiments also support that the VDMs have the capability to filter out sharp minima.
	
	\item  Popular CNNs enhanced by VDMs are compared with the original models on ImageNet, CIFAR-$10$, and CIFAR-$100$. The results are surprising---all the involved models get significant improvements, such as $1.46\%$ top-$1$ test accuracy improvement for ResNet-$20$ on CIFAR-$100$, $0.43\%$ improvement for PreResNet-$18$ on ImageNet, and $1.19\%$ improvement for EfficientNet-B$0$ on CIFAR-$100$. These improvements are significant for the involved image classification tasks, especially without introducing additional tricks.
\end{itemize}

\section{Related work}\label{rw}

%In this section, previous work related to this paper, including visualization of the loss surface and the relationship between the sharpness of the loss surface and the generalization performance, are presented.%%
%\subsection{Visualization of the loss surface}
%\subsection{Relationship between sharpness and generalization}
%The training loss obtained by different settings may be similar, but the test loss could be significantly different \cite{Hochreiter1997Flat}.
The connection between the flatness of a minimum and generalization is first investigated in \cite{Hochreiter1997Flat}, with a method of searching the flat minimum. A visualization method is provided by Li \textit{et al.} to illustrate the potential relationship between hyperparameters and generalization ability of CNNs \cite{li2018visualizing}. This visualizing method gives us an intuitive picture of loss surfaces to inspire the studies of deep learning. It is observed in \cite{li2018visualizing} and \cite{keskar2016large} that the model with smaller batch performs better than that with a larger batch in generalization, and the minima obtained by small-batch methods are flat. In addition, the $(\mathcal{C}_\epsilon, A)$-sharpness to measure the sharpness at a specified point is defined in \cite{keskar2016large}, and the experimental results support that the generalization performance is highly related to the sharpness. However, research in \cite{dinh2017sharp} introduces a re-parametrization method and indicates that the previous measurements of sharpness are not sufficient to predict the generalization capability. Moreover, it is shown in \cite{wang2018identifying} that the generalization ability is affected by the scale of samples, Hessian matrix with its Lipschitz constant of the loss function, and the number of parameters. Additionally, the flat minimum occupies a larger parameter space than the sharp minimum \cite{huang2019understanding}. Specifically, the possibility that the critical point is obtained is positively correlated with the volume of the region in the high-dimensional space rather than the width in a single dimension. To eliminate sharp minima, SmoothOut is proposed in \cite{wen2018Smoothout}. Both theoretical and practical evidence shows that SmoothOut can find flat minima. However, SoothOut still requires the computation of noise injecting and de-noise process. Compared with SmoothOut, the proposed deformation-function-based method in this paper only brings the computation overhead of the VDM acting on the loss function. In practice, there is no significant time consumption difference between the CNN equipped with VDM and the original CNN (see supplementary material).
Overall, these studies demonstrate the importance of sharpness around the minima.
\section{Method}\label{method}
In this section, the method of deforming the loss surface is proposed, followed by theoretical analyses of flat minima filter.

\begin{figure*}[htbp]
	\centering
	\subfigure[AP VDM]{
		\label{d_delta_sub_a}
		\includegraphics[width=2.15in]{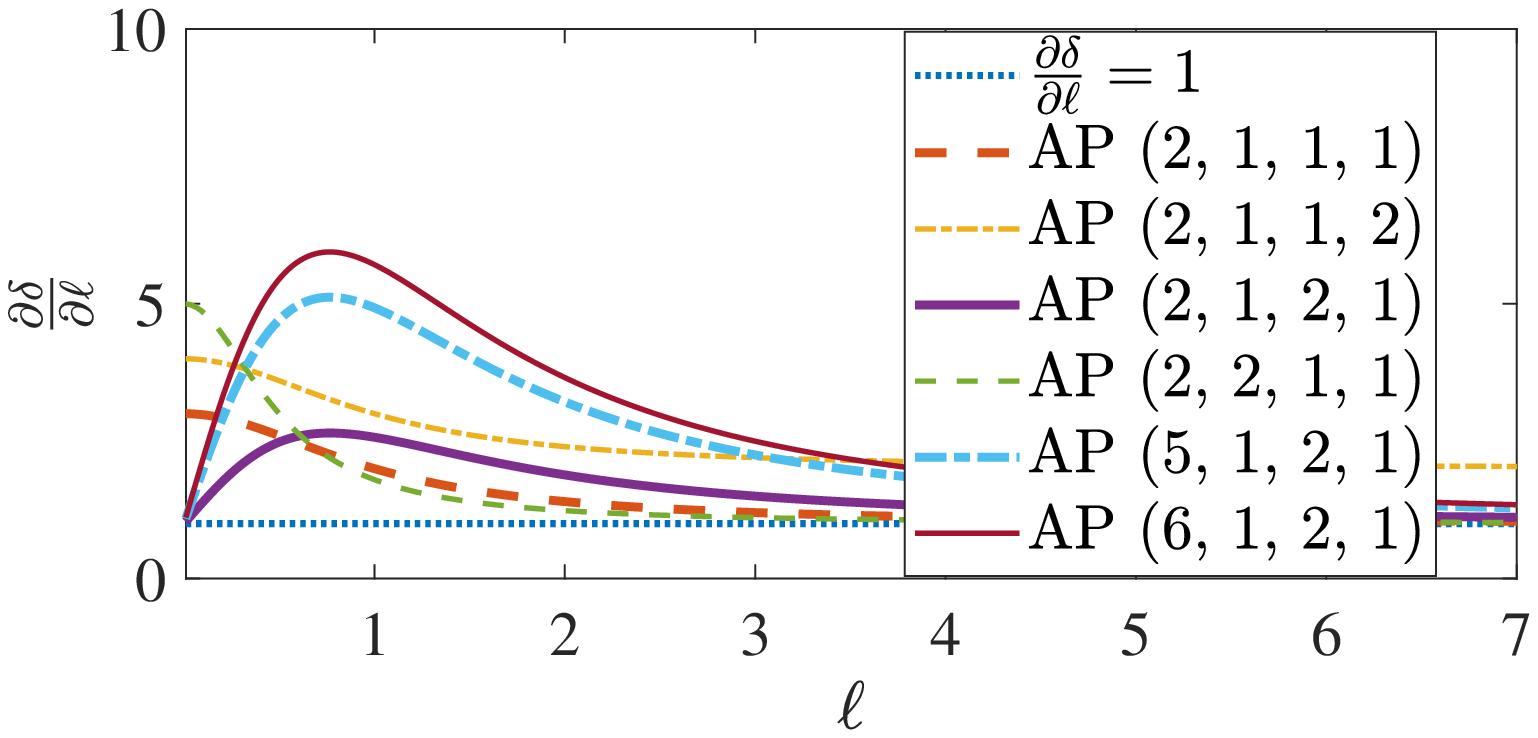}
	}
	\subfigure[LE VDM]{
		\label{d_delta_sub_b}
		\includegraphics[width=2.15in]{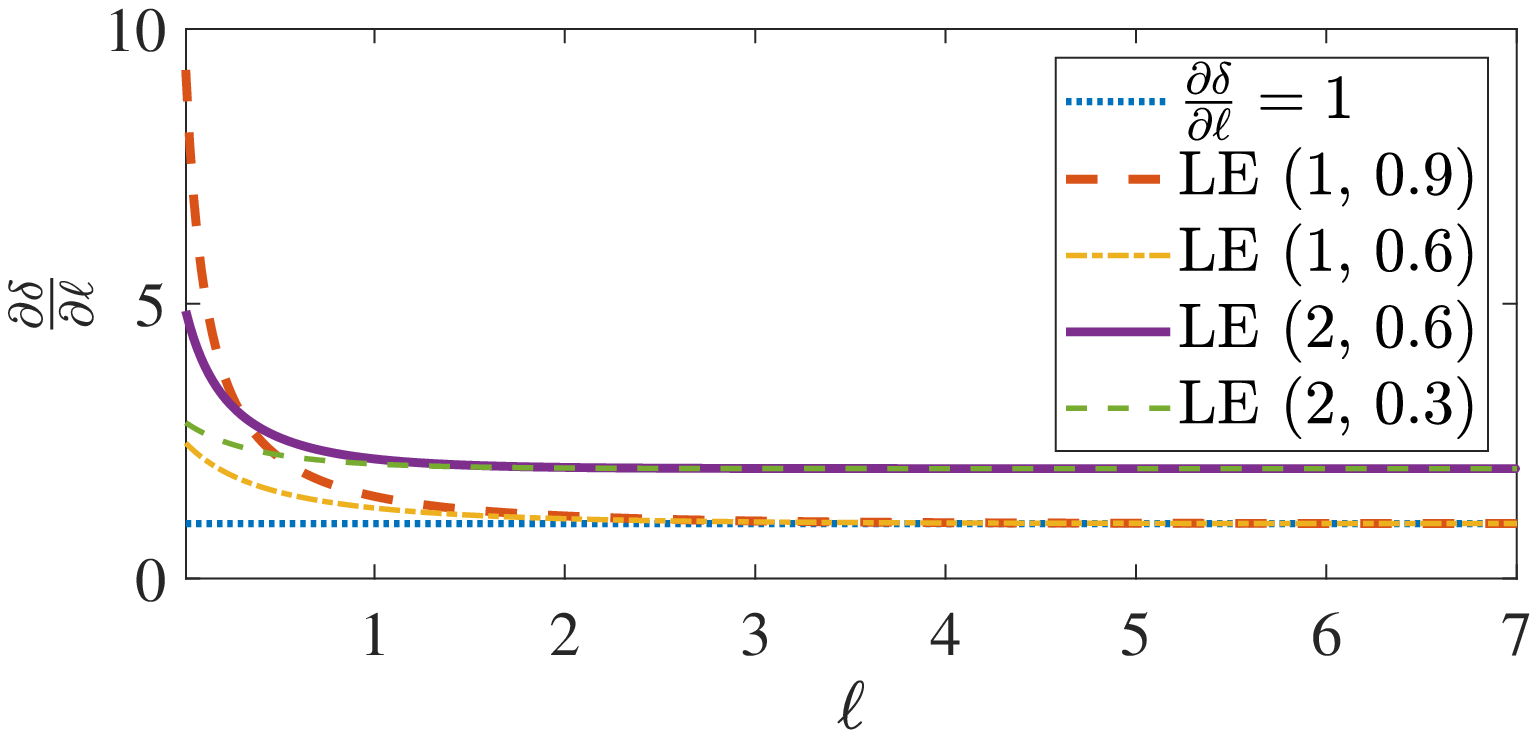}
	}
	\subfigure[]{
	\label{d_delta_sub_c}
	\includegraphics[width=2.15in]{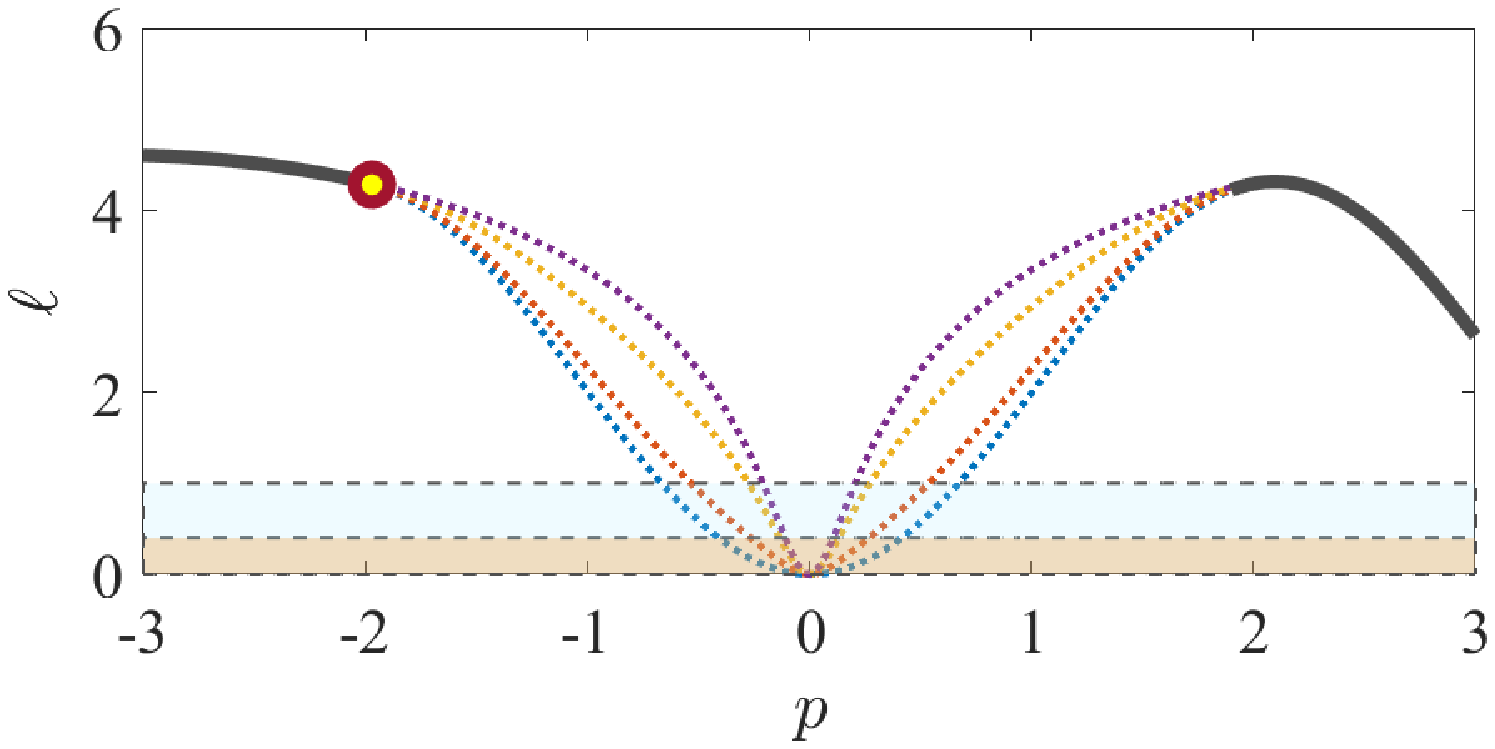}
}
	\caption{Derivatives of (a) AP and (b) LE VDMs, and an illustration (c) about the influence of loss value on the relationship between steepness and flatness.}
	\label{d_delta}
\end{figure*}

\subsection{Deforming the loss surface}\label{Def}

For deforming the loss surface so as to obtain ideal parameters, deformation mapping $\bigvee$ is proposed.
\begin{definition}
Assume that $\boldsymbol{p} = [p_1, p_2, \cdots, p_{\tilde{n}}]^{\top} \in \mathbb{R}^{\tilde{n}}, \check{\boldsymbol{p}} = [\check{p}_1, \check{p}_2, \cdots, \check{p}_{\tilde{n}}]^{\top} \in \mathbb{R}^{\tilde{n}}, \ell(\boldsymbol{p}) \in \mathbb{R}, \check{\ell}(\check{\boldsymbol{p}}) \in \mathbb{R}$, $\bigvee: \mathbb{R}^{\tilde{n}+1} \rightarrow \mathbb{R}^{\tilde{n}+1}$ is an element-wise mapping (i.e. ${\ell} \rightarrow \check{\ell}, {p}_n \rightarrow \check{p}_n$, where $n = 1, 2, \cdots, \tilde{n}$ represent the $n$th element of $\boldsymbol{p}$), and the set $\{\check{\ell}, \check{p}_1, \check{p}_2, \cdots, \check{p}_{\tilde{n}}\} = \bigvee (\{ \ell,  p_1, p_2, \cdots, p_{\tilde{n}}\})$, where $\tilde{n}$ denotes the dimension of the parameter vector, $\cdot^{\top}$ denotes the transposition of a vector, $ \bigvee (\{ \ell,  p_1, p_2, \cdots, p_{\tilde{n}}\})$ means $ \bigvee$ act on set $\{ \ell,  p_1, p_2, \cdots, p_{\tilde{n}}\} $. For an optimization problem $\min_{\boldsymbol{p}}~\ell(\boldsymbol{p})$, $\bigvee$ is called a deformation mapping if the following conditions are met:
\begin{itemize}
\item $\forall \check{\boldsymbol{p}}, \nabla_{\check{\boldsymbol{p}}} \check{\ell}(\check{\boldsymbol{p}}) \exists$ (i.e. for all deformed parameters, $\check{\ell}(\check{\boldsymbol{p}})$ can be partially derived);
%\item $\exists \{ \ell,  \boldsymbol{p}\}, \bigvee (\{ \ell,  \boldsymbol{p}\}) \neq \{ \ell,  \boldsymbol{p}\}$ (i.e. deformation effect exists);
\item $\forall n \in [1, \tilde{n}]$, if $p^{[i]}_n \leq p^{[j]}_n$, then $\check{p}^{[i]}_n \leq \check{p}^{[j]}_n$ (i.e. the quantitative relationship of the parameters in the same dimension remains unchanged before and after the deformation);
%\item If $\ell(\boldsymbol{p}^{[i]}) \leq \ell(\boldsymbol{p}^{[j]})$, then $\check{\ell}(\check{\boldsymbol{p}}^{[i]}) \leq \check{\ell}(\check{\boldsymbol{p}}^{[j]})$ (i.e. the quantitative relationship of the losses remains unchanged before and after the deformation),
\end{itemize}
where $p^{[i]}_n$ and  $p^{[j]}_n$ are two arbitrary parameters of dimension $n$, $\ell(\boldsymbol{p}^{[i]})$.
%  and $\ell(\boldsymbol{p}^{[j]})$ are two arbitrary parameter vectors.
\end{definition}
As one kind of deformation mapping with simple form, the VDM is proposed as follows. 
\begin{definition}
	Suppose that $\delta$ is a deformation mapping. 
	For an optimization problem $\min_{\boldsymbol{p}}~\ell$ with $\boldsymbol{p} \in \mathbb{R}^{\tilde{n}}$, $\delta (\{ \ell,  \boldsymbol{p}\})$ is called a vertical deformation mapping (VDM) if the following conditions are met:
	\begin{itemize}
		\item $\delta$ is a deformation mapping;
		\item $\forall n \in [1, \tilde{n}], \check{p}_{n} = p_{n}$;
		\item $\forall \ell \geq 0$, $\frac{\partial \delta}{\partial \ell} \exists$ and $\frac{\partial \delta}{\partial \ell} \geq 0$.
	\end{itemize}
\end{definition}
In the following, on the premise of not causing confusion, we also use $\ell$ to represents the value of $\ell(\boldsymbol{p})$, and use $\delta$ to represents the value of  $\check{\ell}$. 
\begin{theorem}[VDM affect the spectrum of the Hessian]
	\label{1st_2nd}
	Assuming that $\ell: \mathbb{R}^{\tilde{n}} \rightarrow \mathbb{R} $, ${\partial^2 \ell}/({\partial p_r \partial p_s})$ exists and is continuous for any $r, s \in [1, \tilde{n}]$, of which $p_r$ and $p_s$ denote the $r$th and $s$th element of $\boldsymbol{p}$, respectively, and $\delta$ is the VDM, then $\lambda(\check{\mathbf{H}}) = ({\partial \delta}/{\partial \ell} )\lambda(\mathbf{H})$, where $\lambda(\cdot)$ denotes the spectrum of a matrix, $\mathbf{H}$ denotes the Hessian matrix of $\ell$, and $\check{\mathbf{H}}$ denotes the Hessian matrix of $\check{\ell}$.
\end{theorem}

Proof is given in the supplementary material. Besides, one can derive that
$\nabla \delta(\ell(\mathbf{p}))= ({\partial \delta}/{\partial \ell}) \nabla \ell(\mathbf{p}).$
Thus ${\partial \delta}/{\partial \ell}$ can be deemed as a factor that controls the strategy of deformation. If ${\partial \delta}/{\partial \ell}>1$, the loss surface is deformed to be sharper and steeper; if ${\partial \delta}/{\partial \ell}<1$, the surface becomes flatter. 

The algorithm for VDM-enhanced training process is shown in Algorithm \ref{algor1}. It can be seen from Algorithm \ref{algor1} that, deforming the loss surface by VDM can be regarded as acting a function on the loss, and thus introduces little additional computational overhead---${O}(1)$. The training time comparisons are provided in the supplementary material. Moreover, it is worth noting that although the potential parameters that can give the minimum remain the same after deforming the loss surface, the actual parameters obtained by the optimizers with or without deformation could be diverse. The goal of deforming the loss surface in this paper is to affect the trajectory of the optimizer and then find the minimum that is flat on the original loss surface. 
%As discussed in Sections \ref{FMS} and \ref{VisualExp}, deforming the loss surface to be sharper when $\ell$ is small contributes to search the flat minimum instead. Consider that the flatness is more relevant to the landscape near the minimum (i.e., loss value is small), and thus the loss surface where the loss value is small is usually deformed to be sharper in order to filter out sharp minima.

We assume that where the loss value is large, the steepness of the loss surface (represented by the absolute value of the partial derivative) has little correlation with the flatness (represented by the Hessian spectrum or other indicators); where the loss is small, the steeper the loss surface, the flatter it is. We illustrate this point visually by showing multiple loss curves (all indicated by dashed lines) bifurcated by one loss curve (represented by a solid line) in Fig. \ref{d_delta_sub_c}. The relationship between the steepness and the flatness of the extreme point is not close in the high loss area, but close when the loss is small. When $\ell>4$, the steepness of each loss curve is the same, but the sharpness of the extreme value is completely different. However, when $\ell<1$ (blue and orange areas in Fig. \ref{d_delta_sub_c}), the relationship between steepness and sharpness becomes close: the greater the steepness, the sharper the minimum. With this in mind, we make the loss surface to be steeper only when the loss value is not too large. On the other hand, considering that the optimizer needs to ``slow down'' to converge after escaping from the sharp extreme value, sometimes we will also make those areas where the loss value is very close to $0$ (orange areas in Fig. \ref{d_delta_sub_c}) become closer to the original steepness. 

In the following, the method to obtain VDMs is given, and several well-performing VDMs in practice are analysed. Theorem \ref{1st_2nd} shows that ${\partial \delta}/{\partial \ell}$ directly contributes to the spectrum of the Hessian matrix. With this in mind, one can construct $\varphi(\ell) = {\partial \delta}/{\partial \ell}$ firstly, and then use
$\int \varphi(\ell)\mathrm{d}\ell + C$
to obtain the antiderivative of $\varphi(\ell)$, where $C$ is the constant of integration. The VDMs can also be constructed directly rather than by antiderivative.
Next, several well-performing VDMs designed by following the rule discussed previously are given.
The AP VDM is defined as $\check{\ell} = a_1 (\arctan(a_2 \ell))^{a_3}+a_4 \ell$, where $a_1, a_2, a_3, a_4 > 0$ are the coefficients that control the shape of the AP $(a_1, a_2, a_3, a_4)$.
Another well-performing VDM is LE VDM: $\check{\ell} = e_1 \ln(\exp(\ell) - e_2)$, where $e_1, e_2$ are the coefficients that control the shape of the LE $(e_1, e_2)$ with $e_1, e_2 >0$. 
%%%The derivative of LE $(e_1, e_2)$ is
%%% \begin{equation}\label{OutOfNN}
%%% \frac{e_1 \exp(\ell)}{\exp(\ell)-e_2}>e_1.
%%% \end{equation}
Figures \ref{d_delta_sub_a} and \ref{d_delta_sub_b}illustrates the derivatives of the AP and LE VDMs with different coefficients. For example, for LE VDMs with $e_1=1$, the deformed surface remains similar to the original one (i.e. $\partial \delta / \partial \ell = 1$) if $\ell$ is large enough, and the deformed surface to be sharper if $\ell$ approaches zero. 
If $e_2$ takes a value greater than or equal to $1$, it may make $\partial \delta / \partial \ell$ infinite when loss is equal to $0$ or another certain number. However, since the lowest training loss value is not zero in practice (e.g., the lowest training loss for ResNet-$18$ on ImageNet is around $1$), and the learning rate scheduler usually gives a small learning rate in the later stage of training, $e_2>=1$ is also applicable in some cases.

Note that one of the differences between learning rate scheduler and VDM is that the learning rate scheduler is a function of the epoch, while VDM is a function of loss. In order to filter out sharp minima, the degree of steepness of the deformed loss surface depends on the loss value, not the epoch. In addition, VDM can be used with various learning rate scheduler. 
%To compare the performance between learning rate schedulers and VDMs, Table \ref{LRS} are provided (see supplementary material for more comparisons on CIFAR-$100$).

%\begin{table*}[tbp]
%	\centering
%	\caption{Comparisons among learning-rate-scheduler-equipped PreResNet-$20$ and loss-surface-deformed PreResNet-$20$ on CIFAR-$10$.}
%	\begin{tabular}{lllllllllllllll}
%		\toprule
%		\multirow{1}{*} &$1 \eta$ (Step decay)   &$3 \eta$  &$5 \eta$ &$10 \eta$   &Cosine &HTD(-6, 3)   & HTD(-4, 4)&LE $(1, 0.99)$ \\
%		%&LE $(1, 1)$  &LE $(1, 0.95)$ &AP $(5, 1, 1, 1)$\\
%		
%		\midrule
%		\multirow{1}{*} {Acc. (\%)} &$91.88$    &$92.35$   &$92.25$ &$91.52$   &$92.13$ &$92.44$    & $92.30$&$\textbf{92.64}$ \\ %&$92.40$ &$92.09$ &$92.22$\\
%		\bottomrule
%	\end{tabular}
%	\label{LRS}
%\end{table*}

In addition to the proposed VDM acting on $\ell$, some transformations on the parameters can also be regarded as cases of deformation mappings. For example, reparameterization \cite{dinh2017sharp}.
Specifically, suppose that $\sigma: \mathbb{R}^{\tilde{n}} \rightarrow \mathbb{R}^{\tilde{n}}$ is bijective, then the parameter is transformed by $\varrho = \sigma(p)$. This approach can be described intuitively as scaling the loss surface along the coordinate axes corresponding to parameters. 
\cite{dinh2017sharp} discusses a static situation: if the optimizer just stays at the critical point, changing the sharpness by reparameterization does not affect the parameters found by the optimizer. However, we can use reparameterization as an approach to dynamically affect the behaviour of the optimizer during training, e.g., to make the parameters closer to $0$. 
Due to space limitations, the details will not be investigated in this paper.
\begin{algorithm}[tb]
	\setstretch{1.35}
	\caption{Deforming the loss surface by VDM. }
	\label{alg}
	\begin{algorithmic}\label{algor1}
		\REQUIRE
		$h$: A group of hyperparameters includes learning rate $\eta$, and other hyperparameters required by the specific optimization method, such as exponential decay rates $\beta_1$ and $\beta_2$ for \textsc{Adam}, and momentum $m$ for SGD with momentum (SGDM). \\
		\REQUIRE
		$\Theta$: A function with respect to gradients $\boldsymbol{g}^{(1)}, \cdots, \boldsymbol{g}^{(k-1)}$, $h$, and step $k$. The value of $\Theta(\boldsymbol{g}^{(1)}, \cdots, \boldsymbol{g}^{(k-1)}, h, k)$ represents the increment from  $\boldsymbol{p}^{(k-1)}$ to $\boldsymbol{p}^{(k)}$ (i.e. $\triangle \boldsymbol{p}^{(k)}$). 
		In practice, not all of these independent variables will be explicitly involved in parameter updating. For example, in SGDM, $\triangle \boldsymbol{p}^{(k)}$ contains information from $\boldsymbol{g}^{(1)}, \cdots, \boldsymbol{g}^{(k-1)}$, thereby avoiding explicit involvement of  $\boldsymbol{g}^{(1)}, \cdots, \boldsymbol{g}^{(k-1)}$. \\
		\REQUIRE
		$\boldsymbol{p}_0$: Initial value of the parameter $\boldsymbol{p}$ to be updated.\\
		\REQUIRE
		$\ell$: The loss function.\\
		\REQUIRE
		${\delta}$: The VDM.\\
		%\REQUIRE
		%$\phi$: $\phi = \phi(e, \Theta)$ is a non-linear mapping from $e$ to $\phi$, where $\Theta$ is the update factor.\\
		$k \leftarrow 0$
		\WHILE{$\ell$ is not small enough}
		\STATE{
			$k \leftarrow k + 1$\\
			%$\Theta=\phi(e, \Theta)$\\
			$\boldsymbol{g}^{(k-1)} \leftarrow \nabla \delta ((\ell(\boldsymbol{p}))^{(k-1)})$\\
			$\triangle \boldsymbol{p}^{(k)} \leftarrow \Theta(\boldsymbol{g}^{(1)}, \cdots, \boldsymbol{g}^{(k-1)}, h, k)$\\
			$\boldsymbol{p}^{(k)} \leftarrow \boldsymbol{p}^{(k-1)}+ \triangle \boldsymbol{p}^{(k)}$\\
		}
		\ENDWHILE
		\RETURN $\boldsymbol{p}^{(k)}$
	\end{algorithmic}
\end{algorithm}

\subsection{Flat minima filter} \label{FMS}
Although many minima for deep learning models lead to low training error, not all of them ensure generalization \cite{chaudhari2019entropy}.
Specifically, there is a trend that flat minima or minima on the flat side of the asymmetric valleys generalize better than sharp minima \cite{he2019asymmetric, keskar2019large}. It is worth noting that, if the deformed model gives the same parameters with the original one, although the shape of the loss surface is changed, the accuracy remains the same. With this in mind, our goal is to deform the loss surface to affect the trajectories of optimizers, so that on the deformed loss surface and the original loss surface are diverse so that the parameters obtained finally are diverse as well, as shown in Section \ref{VisualExp}. With this in mind, the focus is deforming the loss surface to obtain the well-generalizable minima, while sharp minima are filtered out.
For illustrations, the trajectories of GD with momentum (GDM) on simulated loss surfaces are given to demonstrate that the VDM can play a role as a flat minima filter, which makes the optimizer to escape from the sharp minimum, as depicted in Fig. \ref{filter_sim}.

In the following, the minima obtained by VDM are investigated theoretically. Here we make some global assumptions that apply to all the following theoretical discussions: (1) gradient descent (GD) optimizer equipped with the VDM is taken as the optimizer, which follows $\boldsymbol{p}^{(k+1)} = \boldsymbol{p}^{(k)} - \eta \nabla \delta((\ell(\boldsymbol{p}))^{(k)})$; (2) total step $\tilde{k}$ is large enough, the lower bound of the original loss is 0; (3) the smaller the absolute value of the gradient near the critical point (excluding the critical point), the flatter the loss surface; (4) the closer to the critical point, the stronger the correlation between gradient and flatness. We first discuss the one-dimensional case, and then the high-dimensional case. 
%%%The parameters in each step form a sequence $(\boldsymbol{p}^{(1)}, \cdots, \boldsymbol{p}^{(k)}, \cdots, \boldsymbol{p}^{(\tilde{k})})$, and the projection of the sequence onto direction $\boldsymbol{q}_n$ is $(\boldsymbol{p}^{(1)}_n, \cdots, \boldsymbol{p}^{(k)}_n, \cdots, \boldsymbol{p}^{(\tilde{k})}_n)$, where the $i$th element $q_{n, i}$ of $\boldsymbol{q}_n$ is defined as: $q_{n, i} = 0$, if $i \neq n$; $q_{n, i} = 1$, if $i = n$.

\begin{definition}\label{MRegion}
	Assume that within an interval $(x^{[a]}, x^{[b]})$, $f(x) : \mathbb{R} \rightarrow \mathbb{R}$ is second-order differentiable, there are only three extreme points
	$(x^\mathrm{[min]}$, $f(x^\mathrm{[min]}))$, $(x^\mathrm{[max_1]}, f(x^\mathrm{[max_1]}))$, and
	$(x^\mathrm{[max_2]}, f(x^\mathrm{[max_2]}))$ with
	$(\partial^2 f(x)/ \partial (x)^2)^\mathrm{[min]}>0$, $(\partial f(x) / \partial (x)^2)^\mathrm{[max_1]}<0$, $(\partial f(x) / \partial (x)^2)^\mathrm{[max_2]}<0$, and $x^\mathrm{[max_1]}<x^\mathrm{[max_2]}$, then $(x^{[a]}, x^{[b]})$ is called local concave region (or ``local M region'' in the sight of geometrical shape) for $f(x)$, interval $(x^\mathrm{[max_1]},  x^\mathrm{[max_2]})$ is called a ``V part'' of a local M region.
\end{definition}

\begin{theorem}\label{th_dp_bound}
	%[VDM contributes to the Lipschitz constant]
	Follow the notations for definition \ref{MRegion}, assume that an interval $(p^{[a]}, p^{[b]})$ is a local M region for $\ell(p)$ with extreme points $(p^\mathrm{[min]}$, $\ell(p^\mathrm{[min]}))$, $(p^\mathrm{[max_1]}, \ell(p^\mathrm{[max_1]}))$, and $(p^\mathrm{[max_2]}, \ell(p^\mathrm{[max_2]}))$, where $p^\mathrm{[max_1]} < p^\mathrm{[max_2]}$, $\ell(p): \mathbb{R} \rightarrow \mathbb{R}$ is second-order differentiable, initial position is within $(p^\mathrm{[max_1]},  p^\mathrm{[max_2]})$,  and gradient descent is taken as the optimizer.
	Let $\triangle p^\mathrm{[max]} = p^\mathrm{[max_2]} - p^\mathrm{[max_1]}$. For a sufficiently large $\tilde{k}$, if $p^{(k+1)}$ remains in the V part $(p^{[max_1]}, p^{[max_2]})$, $\partial (\ell(p))^{(k)} / \partial p^{(k)}$ is bounded by
	\begin{equation}\label{bounded}
	\bigg| \frac{\partial (\ell(p))^{(k)}} {\partial p^{(k)}}\bigg| \leq
	\frac{\triangle p^\mathrm{[max]}}{\eta \frac{\partial \delta}{\partial \ell} }.
	\end{equation}
\end{theorem}
Proof is given in the supplementary material.
%\begin{lemma}\label{Lp_lemma}
%	Under assumptions of Theorem \ref{th_dp_bound}, Lipschitz condition is met:
%	\begin{equation}\label{Lp_1}
%	||(\ell(p))^{(k)} - \ell(\hat{p}^{(k)}) ||  \leq
%	\frac{\triangle p^\mathrm{[max]}}{\eta  \frac{\partial \delta}{\partial \ell}  }  { || p^{(k)}-\hat{p}^{(k)}|| },
%	\end{equation}
%	where $\hat{p}^{(k)} \in \mathbb{R} $ is arbitrary one-dimensional parameter $p$ on step $k$.
%\end{lemma}

\begin{figure}
	\subfigure[3D view]{
%		\psfrag{$\delta$}{$\check{\ell}$}
		\includegraphics[width=1.571in]{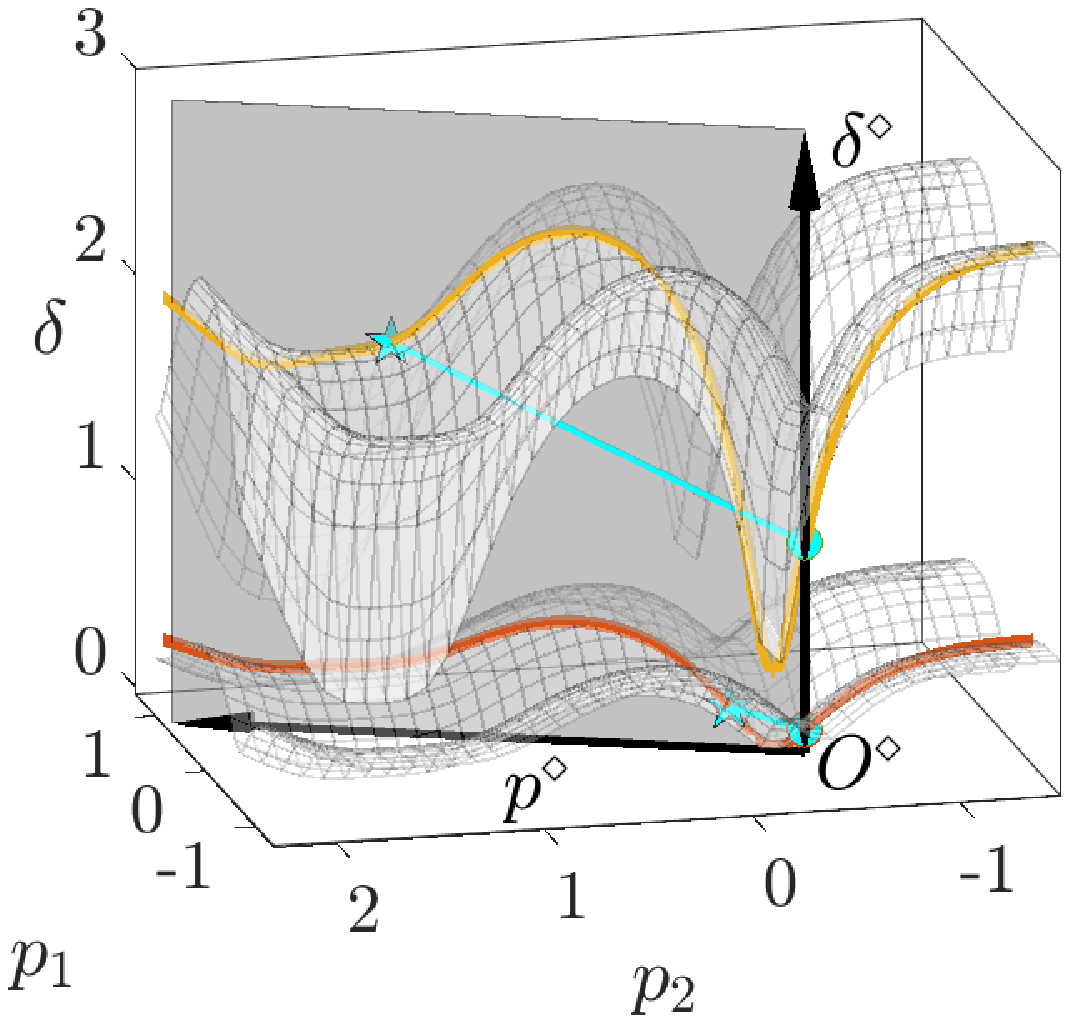}
		\label{fig:highdims_a}
	}
	\subfigure[Front view]{
		\includegraphics[width=1.571in]{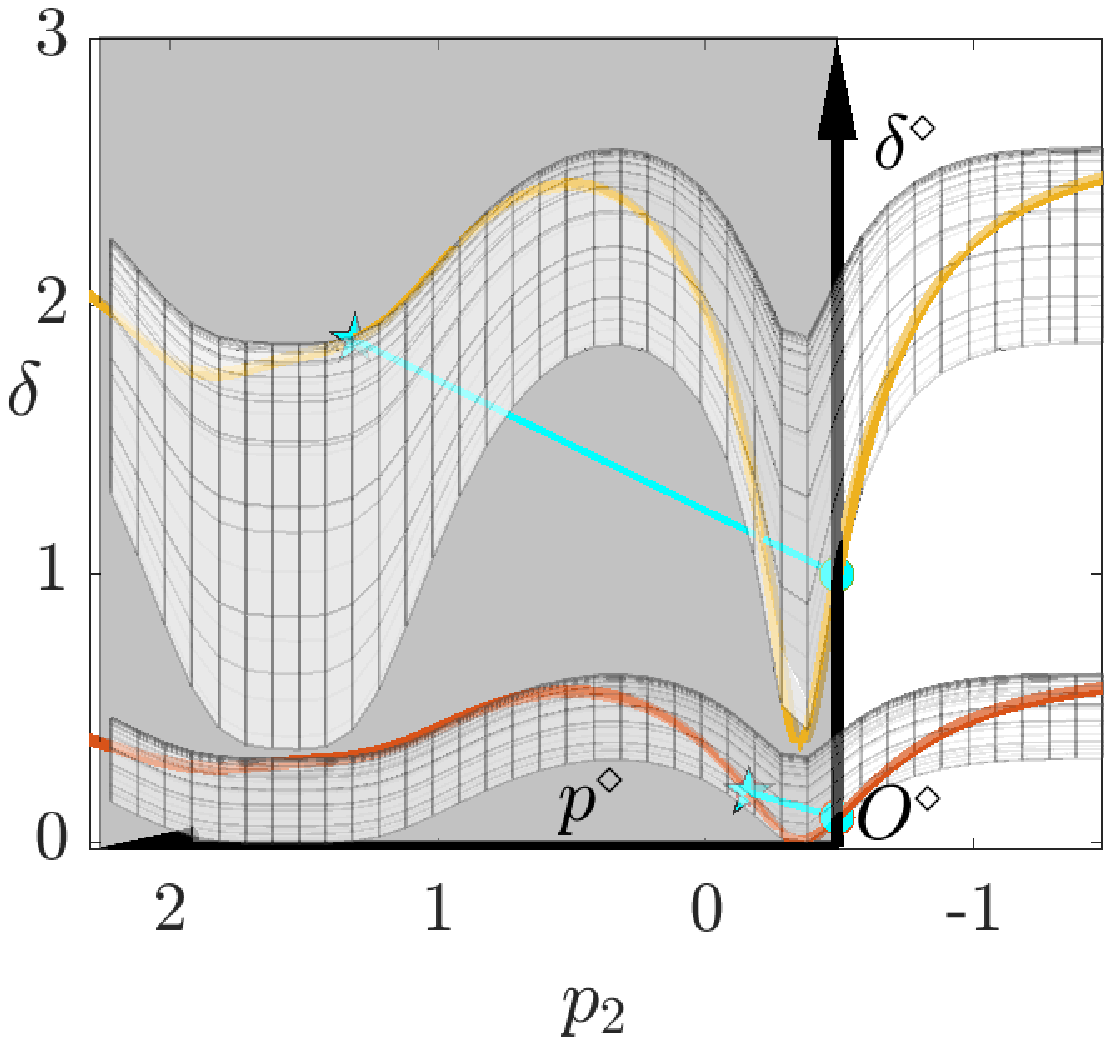}
		\label{fig:highdims_b}
	}
	\caption{Illustration for high dimensional cases. We consider only one more step from the current location for high dimensional cases. Best viewed in color. }
	\label{HD}
\end{figure}

\begin{remark}
	Assume that in the M region, except for the V part, the optimizer does not encounter those positions where ${\partial (\ell(p))^{(k)}} / {\partial p^{(k)}}$ tends to zero. Under this assumption, we can know that once the optimizer locates out of the V part, it will leave the M region within limited steps. With this in mind, if $\forall k \leq \tilde{k}$, ${p}^{(k)}\in (p^{[a]}, p^{[b]})$, i.e. the optimizer remains in the M region, then we have a conclusion that the optimizer does not encounter positions with large gradients. Moreover, ${p}^{(\breve{k})}$ can be deemed as a set sampled from a subspace of the loss surface where $\breve{k} = \{k~|~{p}^{(k)}\in (p^{[a]}, p^{[b]})\}$. Assume the samples contain some information about the nature of the region. If the optimizer does not encounter positions with large gradients within an interval for all the steps, very likely, the gradients in that interval are not so large. In addition, from inequality \eqref{bounded}, together with the assumption that the gradient's absolute value of the low-loss region is positively correlated with the sharpness, one can deduce that the original GD have the capability to filter out sharp minima to some extent. For GD equipped with VDM, if $\partial \delta/\partial \ell$ is increased, the boundary of gradient decreases. This result demonstrates that the deformation method with a larger $\partial \delta/\partial \ell$ tends to give flatter minimum. 
\end{remark}
%\begin{remark}
	For illustrating the high-dimensional case, we provide Fig. \ref{HD}. There are two loss surfaces in Fig. \ref{HD}, in which the deformed loss surface is translated above the original loss surface for clear presentation (this translation does not affect the optimization process). The single-step trajectories on the deformed loss surface and the original loss surface are both in a same plane (discussed formally below), i.e. plane $p^{\diamond} \text{-}~O^{\diamond} \text{-}~ \delta^{{\diamond}}$. We establish a Cartesian coordinate system in this plane, and on this basis, we can define the M region. As can be seen from the plane, the deformed version is more helpful for escaping from the sharp minimum. Next, we formally explain why the single-step trajectories are in the same plane, and how to establish a Cartesian coordinate system for the high-dimensional case. 
	 Consider an arbitrary VDM $\delta$.
	%, and denote the change of the parameter vectors for the original loss surface and deformed loss surface from step $k$ to step $k+1$ as $\triangle \boldsymbol{p}^{(k)} = \boldsymbol{p}^{(k+1)} - \boldsymbol{p}^{(k)} = - \eta \nabla (\ell(\boldsymbol{p}))^{(k)}$ and $\triangle \hat{\boldsymbol{p}}^{(k)} = \hat{\boldsymbol{p}}^{(k+1)} - \boldsymbol{p}^{(k)} = - \eta (\partial \hat{\delta} / \partial \ell) \nabla (\ell(\boldsymbol{p}))^{(k)}$, respectively. 
	Let $p^{\diamond}$ be a line passing through points $q^{(k)} = (p_1^{(k)}, p_2^{(k)}, \cdots, p_{\tilde{n}}^{(k)}, 0)$ and $q^{(k+1)} = (p_1^{(k+1)}, p_2^{(k+1)}, \cdots, p_{\tilde{n}}^{(k+1)}, 0)$, 
	% (considered $(p_1^{(k)}, p_2^{(k)}, \cdots, p_{\tilde{n}}^{(k)})$ as vectors $\boldsymbol{p}^{(k)}$ and $\boldsymbol{p}^{(k+1)}$, respectively), 
	then $p^{\diamond}$ can be written as 
		$p^{\diamond} =  \{ (1-\alpha)(p_1^{(k)}, p_2^{(k)}, \cdots, p_{\tilde{n}}^{(k)}, 0) + \alpha (p_1^{(k+1)}, p_2^{(k+1)}, \cdots, p_{\tilde{n}}^{(k+1)}, 0)~|~\alpha \in \mathbb{R}\} 
		= \{ (p_1^{(k)} - \alpha \eta (\nabla_1 \ell(\boldsymbol{p}))^{(k)}, p_2^{(k)} - \alpha \eta (\nabla_2 \ell(\boldsymbol{p}))^{(k)}, \cdots, p_{\tilde{n}}^{(k)} - \alpha \eta (\nabla_{\tilde{n}} \ell(\boldsymbol{p}))^{(k)}, 0 )~|~\alpha \in \mathbb{R}\},$ 
		where $\nabla_{{n}}$ represents the partial derivative of $\ell$ to $p_{n}$.
As can be derived that,  point $\check{q}^{(k+1)} = ( p_1^{(k)} - \eta \partial \delta/\partial \ell (\nabla_1 \ell(\boldsymbol{p}))^{(k)}, p_2^{(k)} - \eta \partial \delta/\partial \ell (\nabla_2 \ell(\boldsymbol{p}))^{(k)}, \cdots, p_{\tilde{n}}^{(k)} - \eta \partial \delta/\partial \ell (\nabla_{\tilde{n}} \ell(\boldsymbol{p}))^{(k)}, 0 ) $ belongs to $p^{\diamond}$. In the following, we discuss the establishment of a two dimensions Cartesian coordinate system and then discuss definition of the M region in this system. 
Consider a line ${\delta}^{\diamond} = \{ (1-\beta)(p_1^{(k)}, p_2^{(k)}, \cdots, p_{\tilde{n}}^{(k)}, 0) + \beta (p_1^{(k)}, p_2^{(k)}, \cdots, p_{\tilde{n}}^{(k)}, (\ell(\boldsymbol{p}))^{(k)})~|~\beta \in \mathbb{R}\}$, which is perpendicular to $p^{\diamond}$. 
%We set the the axes $p^{\diamond}$ and $\delta^{\diamond}$ based on $p^{\diamond}$ and ${\delta}^{\diamond}$ respectively:
 Notice that the intersection of $p^{\diamond}$ and ${\delta}^{\diamond}$ is $q^{(k)}$, which is also denoted as $O^{\diamond}$. We set $O^{\diamond}$ as the origin of a new Cartesian coordinate system, the orientations of axes $p^{\diamond}$ and $\delta^{\diamond}$ to be the same with vectors $[\boldsymbol{p}^{(k+1)} - \boldsymbol{p}^{(k)},  0]^{\top}$ and $[0, 0, \cdots, 0, 1]^{\top}$ respectively (both of them are in $  \mathbb{R}^{\tilde{n}+1}$), and the length units of the two axes to be the same as the original Cartesian coordinate system (i.e., the coordinate system used to measure $\boldsymbol{p}$ and $\ell$), then a new Cartesian coordinate system is obtained. Denote the original and deformed loss surfaces as $\gamma_{\text{ori}}$ and $\gamma_{\text{defo}}$ respectively, then $m_{\text{ori}} = \gamma_{\text{ori}} \cap (p^{\diamond} \text{-}~O^{\diamond} \text{-}~ \delta^{{\diamond}})$ and $m_{\text{defo}} = \gamma_{\text{defo}} \cap (p^{\diamond} \text{-}~O^{\diamond} \text{-}~ \delta^{{\diamond}})$ are both curves (illustrated in red color and yellow color respectively in Fig. \ref{HD}). Follow the definition of the local M region for one dimensional case, we can define a local M region and the corresponding V part in $p^{\diamond} \text{-}~O^{\diamond} \text{-}~ \delta^{{\diamond}}$ plane similarly. Assume that for each step, optimizer locates in an M region. Note that for each step, we establish a new Cartesian coordinate system. Then we can apply Theorem \ref{th_dp_bound} in each new Cartesian coordinate system, and find that for each step, compared with the original loss surface, the deformed loss surface is more likely to lead to a flat minimum.
\begin{figure}[ht]
	\centering
	\subfigure[]{
		\label{eig_round_a}
		\includegraphics[width=2.581in]{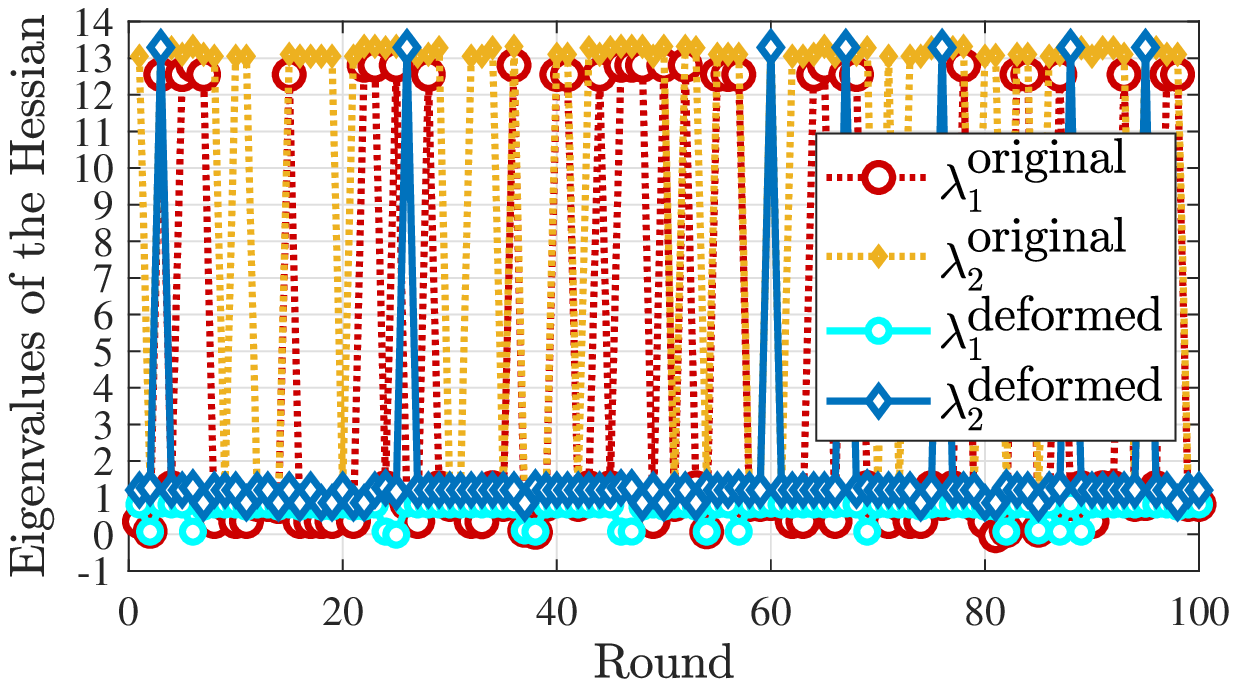}
	}
	\subfigure[]{
		\label{eig_round_b}
		\includegraphics[width=2.551in]{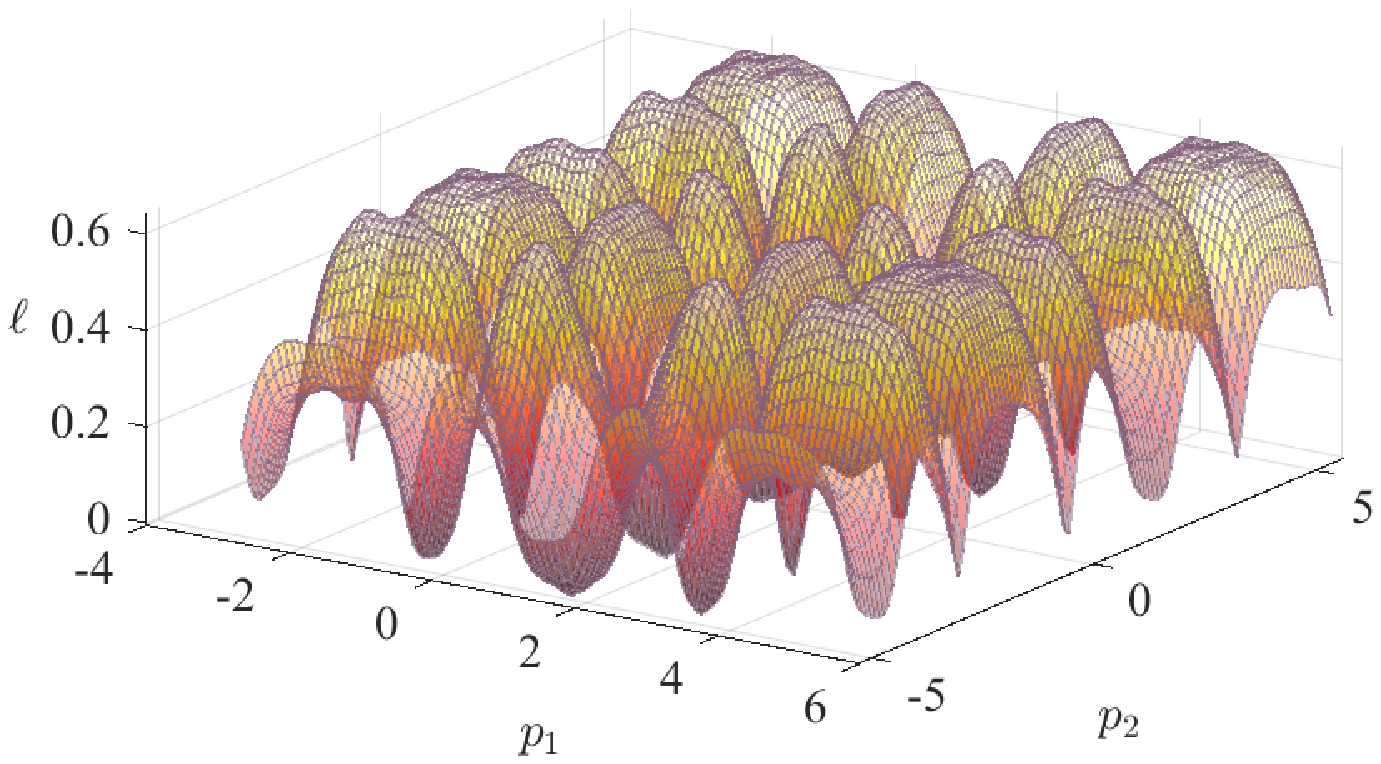}
	}
	\caption{Comparison of eigenvalues of the Hessian for the original loss surface and the deformed one. (a) Eigenvalues of the Hessian in $100$ rounds simulations with $2$-parameters. (b) The landscape of the (original) loss surface. }
	\label{eig_round}
\end{figure}
According to the above analyses, the AP and LE VDM are conducive to search the flat minimum. 
%From Lemma \ref{Lp_lemma}, it can be seen that the VDM is also beneficial for reducing the value of the $(\mathcal{C}_\epsilon, A)$-sharpness defined in \cite{keskar2016large}. 
To give some empirical illustrations, we conduct $2$-parameters simulations on a complex loss surface which contains lots of sharp minima or flat minima (see Fig. \ref{eig_round}). We assign a random initial position to each round of the simulation and perform GD on the simulated loss surface, which contains a large number of sharp minima and flat minima, and then calculate the Hessian eigenvalues of the final positions of the optimizers. Note that the final locations are separately obtained on the original surface or deformed surface, but the eigenvalues are both calculated on the original loss surface. Please see supplementary material for more details about simulation settings.
%loss curves of both the original ResNet-$20$ and the vertical-deformation-mapping-equipped ResNet-$20$ are presented. As shown in Fig. \ref{curve_CE}, AP and LE VDMs can make the loss curve flatter, and the larger $a_1$ or $e_2$ is (it means the loss surface is deformed to be steeper at low-loss region), the flatter the curve is. 
These results demonstrate that the involved VDMs play roles as the flat minimum filters. 
However, if the loss surface becomes too steep, the optimizer may jump to areas with large loss values, and then bring poor performance. 
For example, the loss surface deformed by AP $(10, 1, 2, 1)$ (see Fig. \ref{curve_CE_h}) is flatter than that deformed by AP $(18/\pi, 1, 2, 1)$ (see Fig. \ref{curve_CE_g}), but the training loss in Fig. \ref{curve_CE_h} is higher than the loss in Fig. \ref{curve_CE_g}, and the experimental result is that AP $(18/\pi, 1, 2, 1)$ is better than AP $(10, 1, 2, 1)$. This is a trade-off between the flatness and the (training) loss value. Due to this trade-off and the complexity of the loss surface, we currently cannot find a single VDM that works well in all situations. However, we do get VDMs that can improve the performance of all models in only two function families (i.e., AP and LE).

\section{Experiments}\label{Experiments}

In this section, experiments for various CNNs on ImageNet, CIFAR-$10$, and CIFAR-$100$ are conducted \cite{Krizhevsky09learningmultiple, russakovsky2015imagenet}. The data augmentation methods are random crop and random horizontal flip for both CIFAR ($32 \times 32$) and ImageNet ($224 \times 224$). All models in this paper are trained by SGDM with momentum $0.9$, equipped with Nesterov method and step decay, and the loss function in all the experiments is the cross entropy loss. We did not use additional tricks, such as dropout and additional data augmentation. In order to achieve maximum fairness, we train all comparison models from scratch, by using the same hyperparameters as much as possible. Detailed settings of hyperparameters are provided in the supplementary material.

\subsection{Vertical-deformation-mapping-equipped SGDM find flat minima} \label{VisualExp}
\begin{figure*}[h]
	\centering
	\subfigure[$\check{\ell}= 0.5\ell^2$]{
		\label{curve_CE_a}
		\includegraphics[width=1.6in]{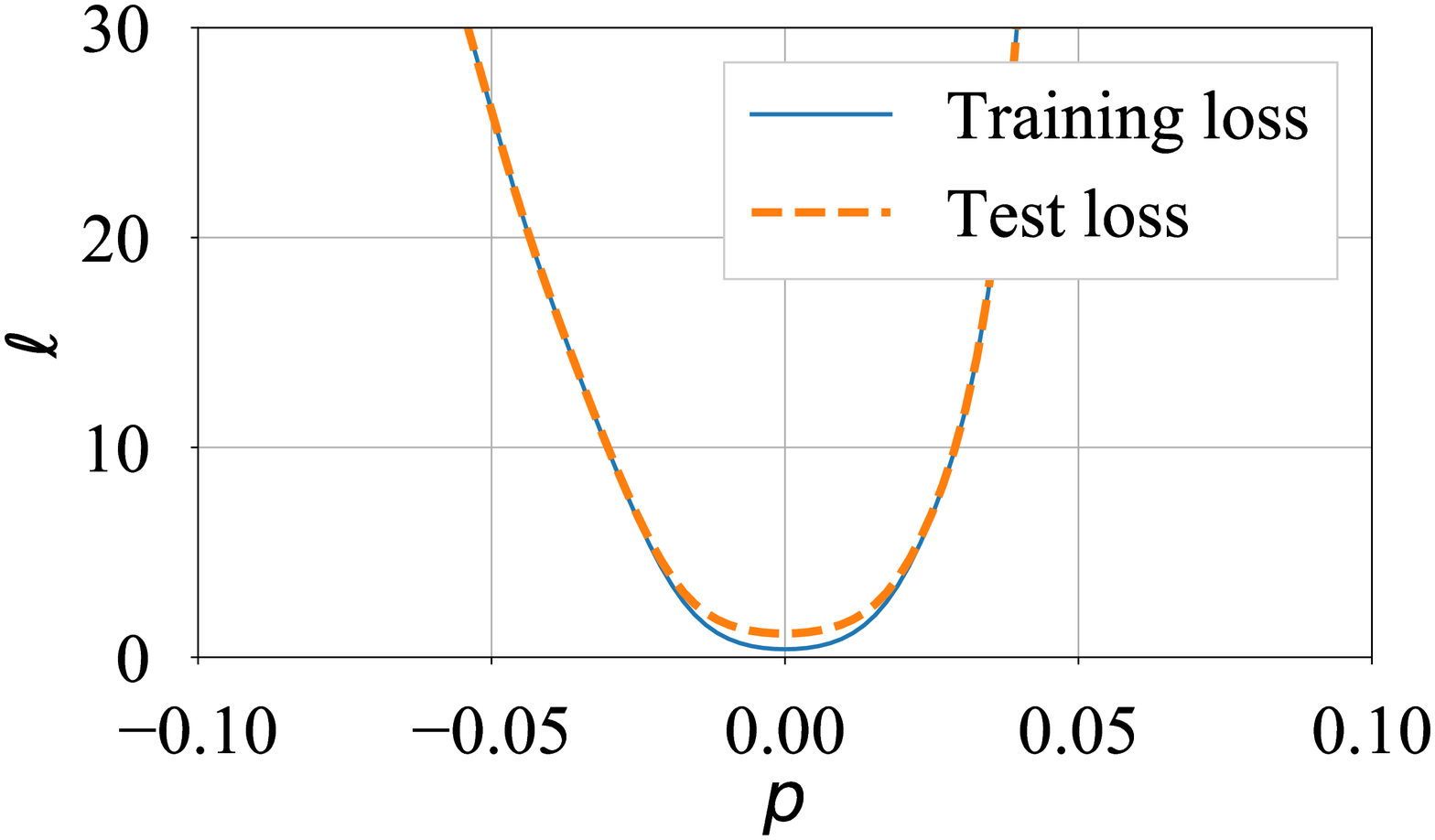}
	}
	\subfigure[The original model]{
		\label{curve_CE_b}
		\includegraphics[width=1.6in]{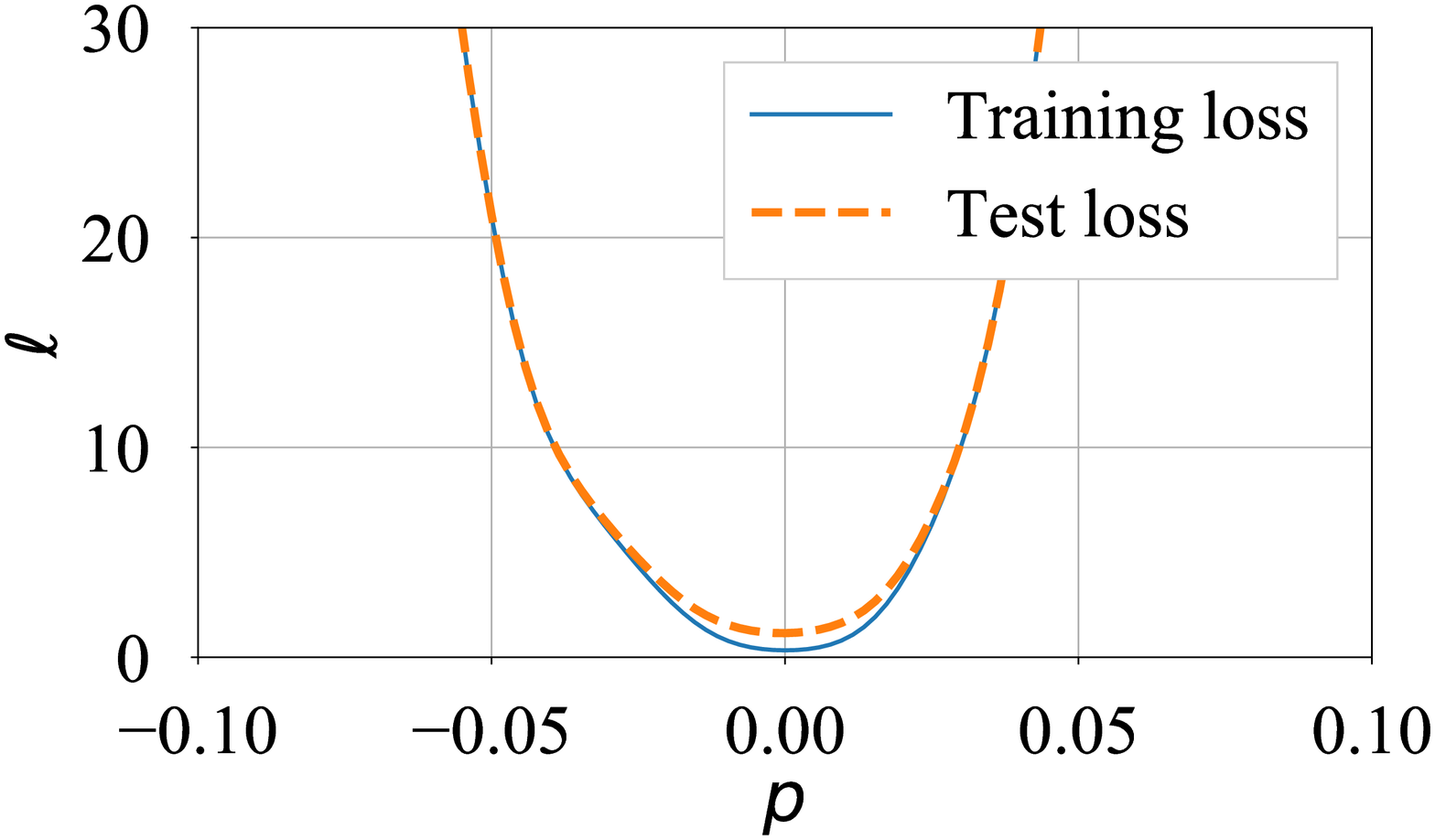}
	}
%	\subfigure[LE $(1, 0.3)$]{
%	\label{curve_CE_h}
%	\includegraphics[width=1.6in]{curve_logexp30.eps}
%}
\subfigure[LE $(1, 0.60)$]{
	\label{curve_CE_c}
	\includegraphics[width=1.6in]{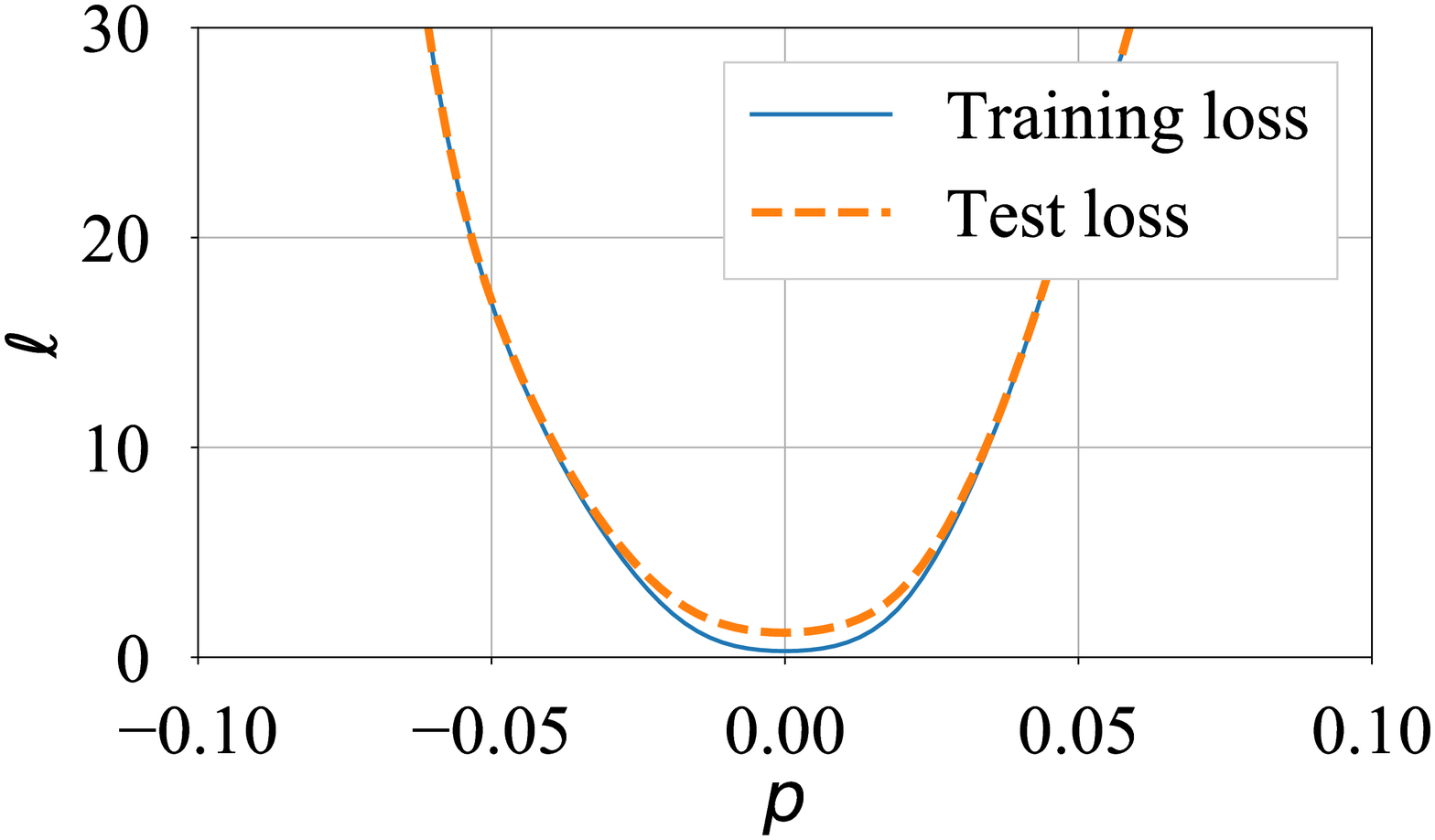}
}
%	\subfigure[LE $(1, 0.8)$]{
%	\label{curve_CE_h}
%	\includegraphics[width=1.6in]{curve_logexp80.eps}
%}
%	\subfigure[LE $(1, 0.85)$]{
%	\label{curve_CE_h}
%	\includegraphics[width=1.6in]{curve_logexp85.eps}
%}
\subfigure[LE $(1, 0.90)$]{
	\label{curve_CE_d}
	\includegraphics[width=1.6in]{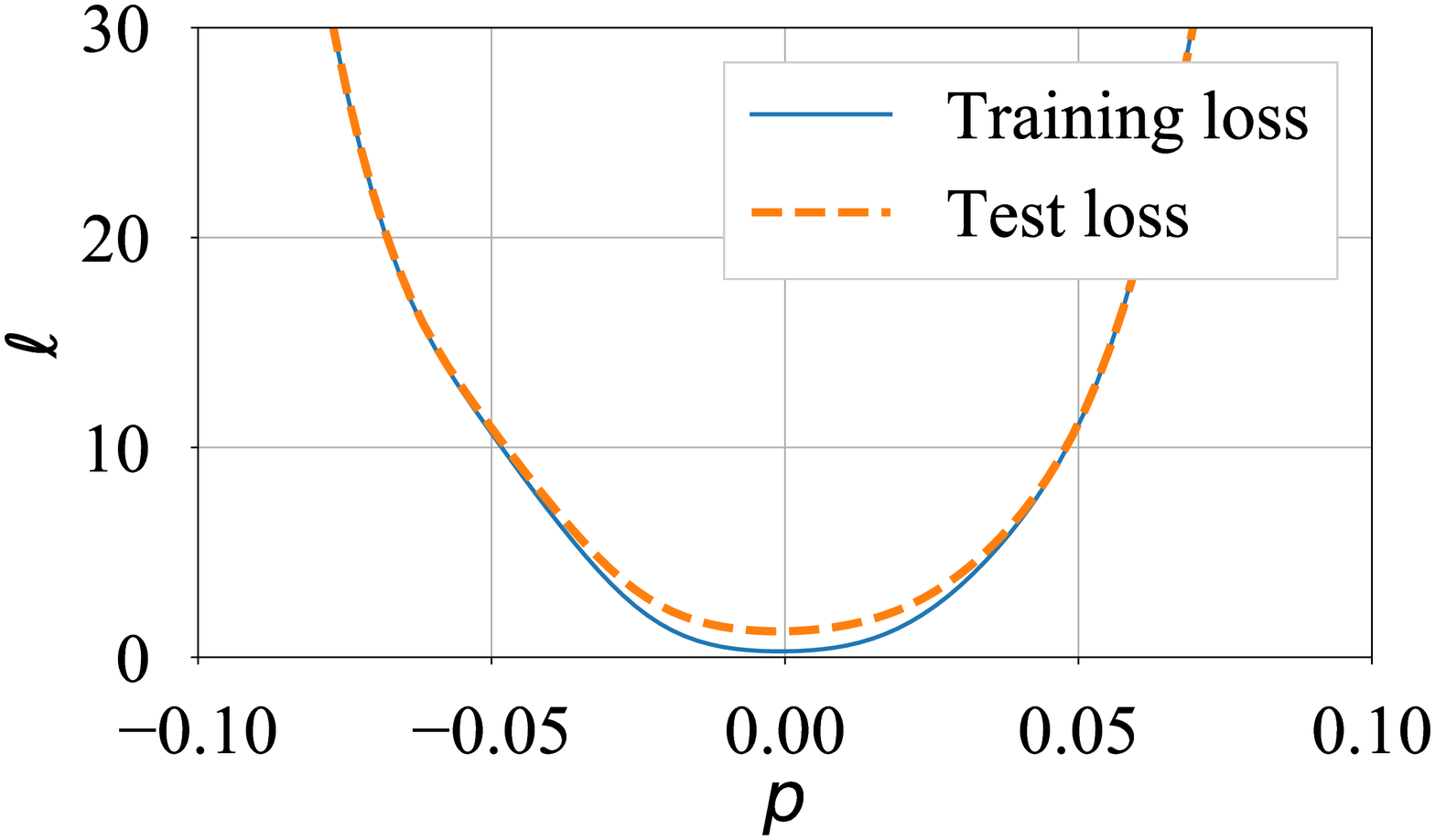}
}
\subfigure[LE $(1, 0.99)$]{
	\label{curve_CE_e}
	\includegraphics[width=1.6in]{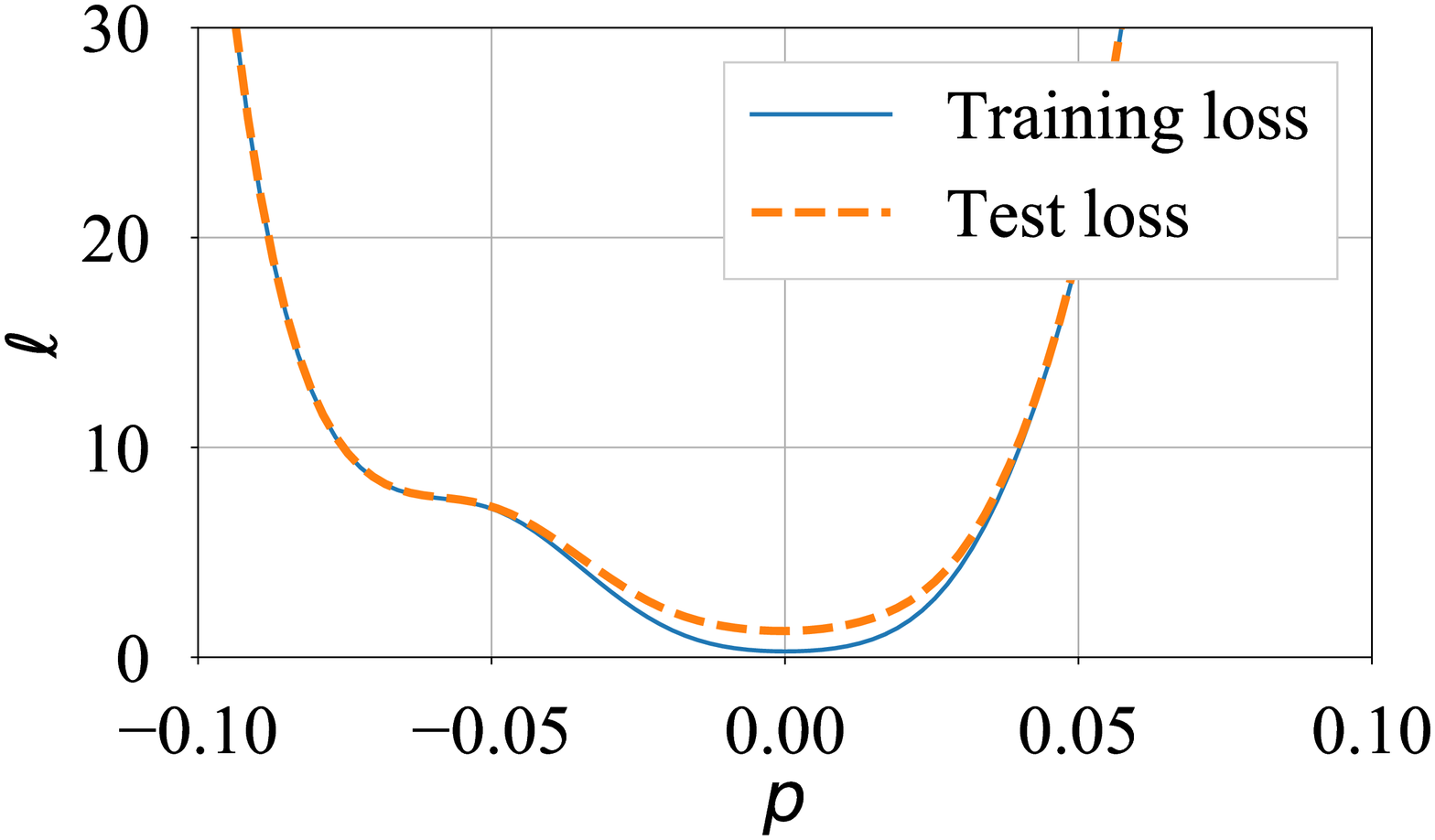}
}
%	\subfigure[AP $(2, 1, 2, 1)$]{
%		\label{curve_CE_f}
%		\includegraphics[width=1.6in]{curve_AP2121.eps}
%	}
	\subfigure[AP $(5, 1, 2, 1)$]{
		\label{curve_CE_f}
		\includegraphics[width=1.6in]{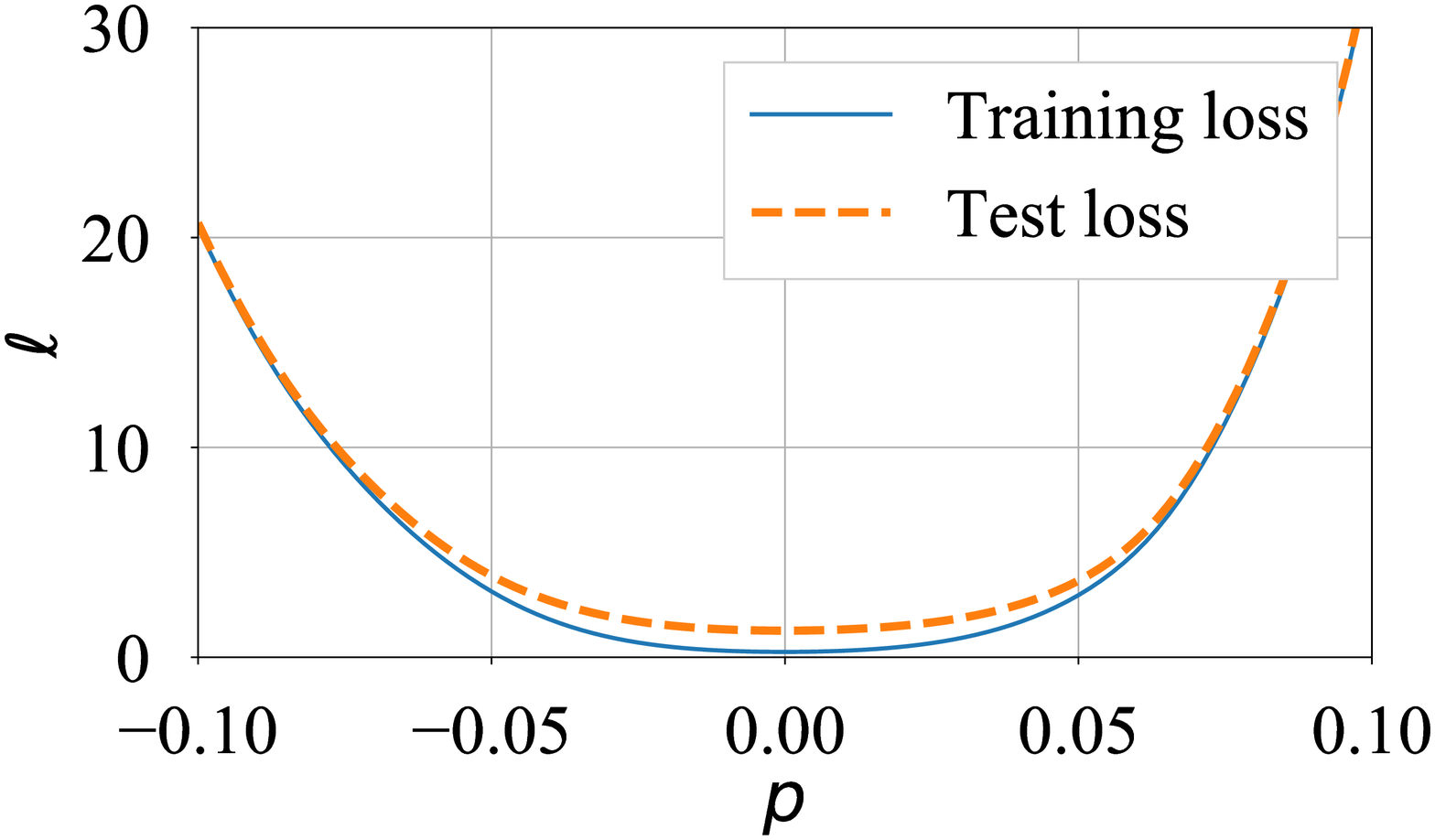}
	}
	\subfigure[AP $(18/\pi, 1, 2, 1)$]{
		\label{curve_CE_g}
		\includegraphics[width=1.6in]{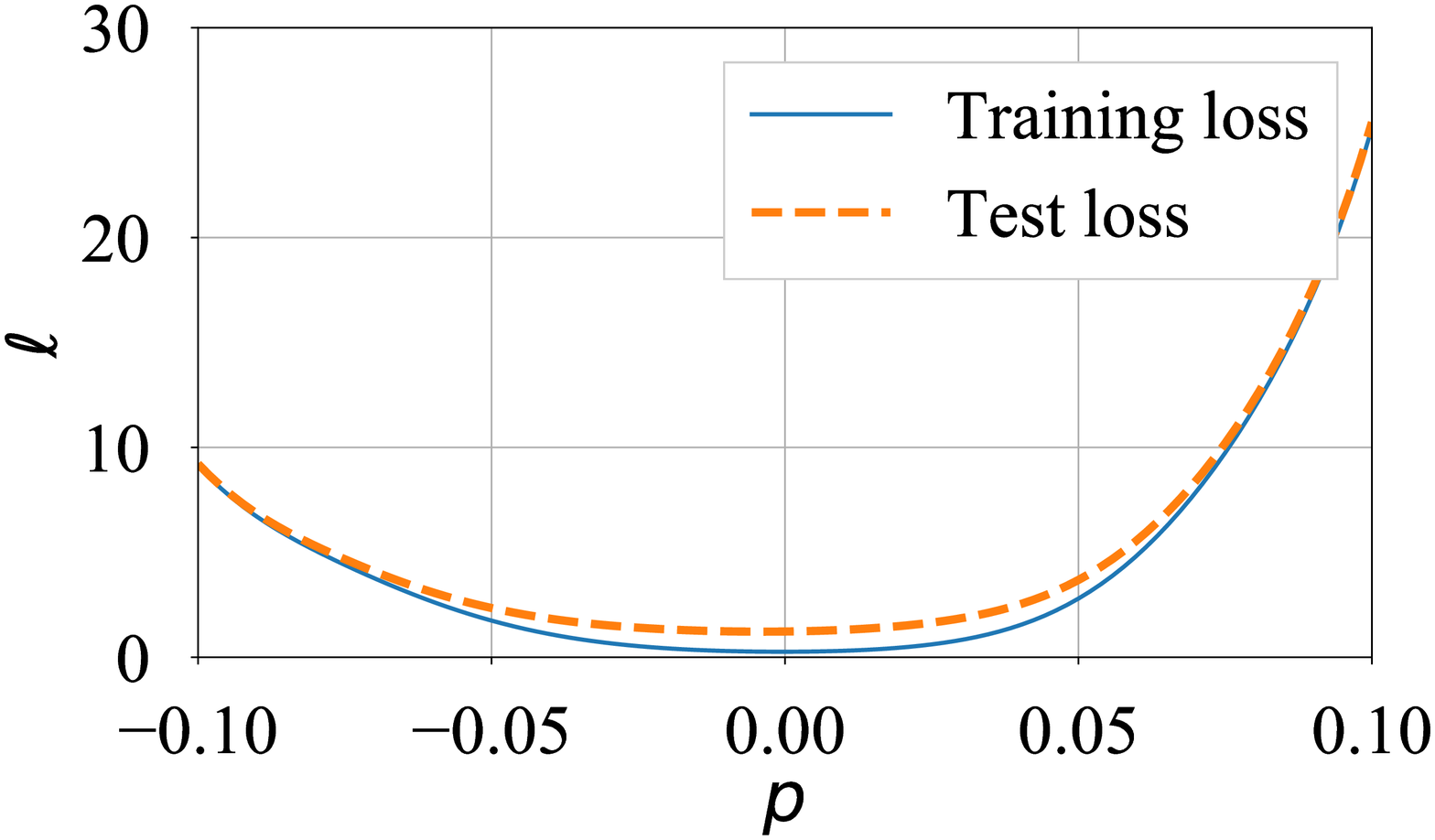}
	}
	\quad
	\subfigure[AP $(10, 1, 2, 1)$]{
		\label{curve_CE_h}
		\includegraphics[width=1.6in]{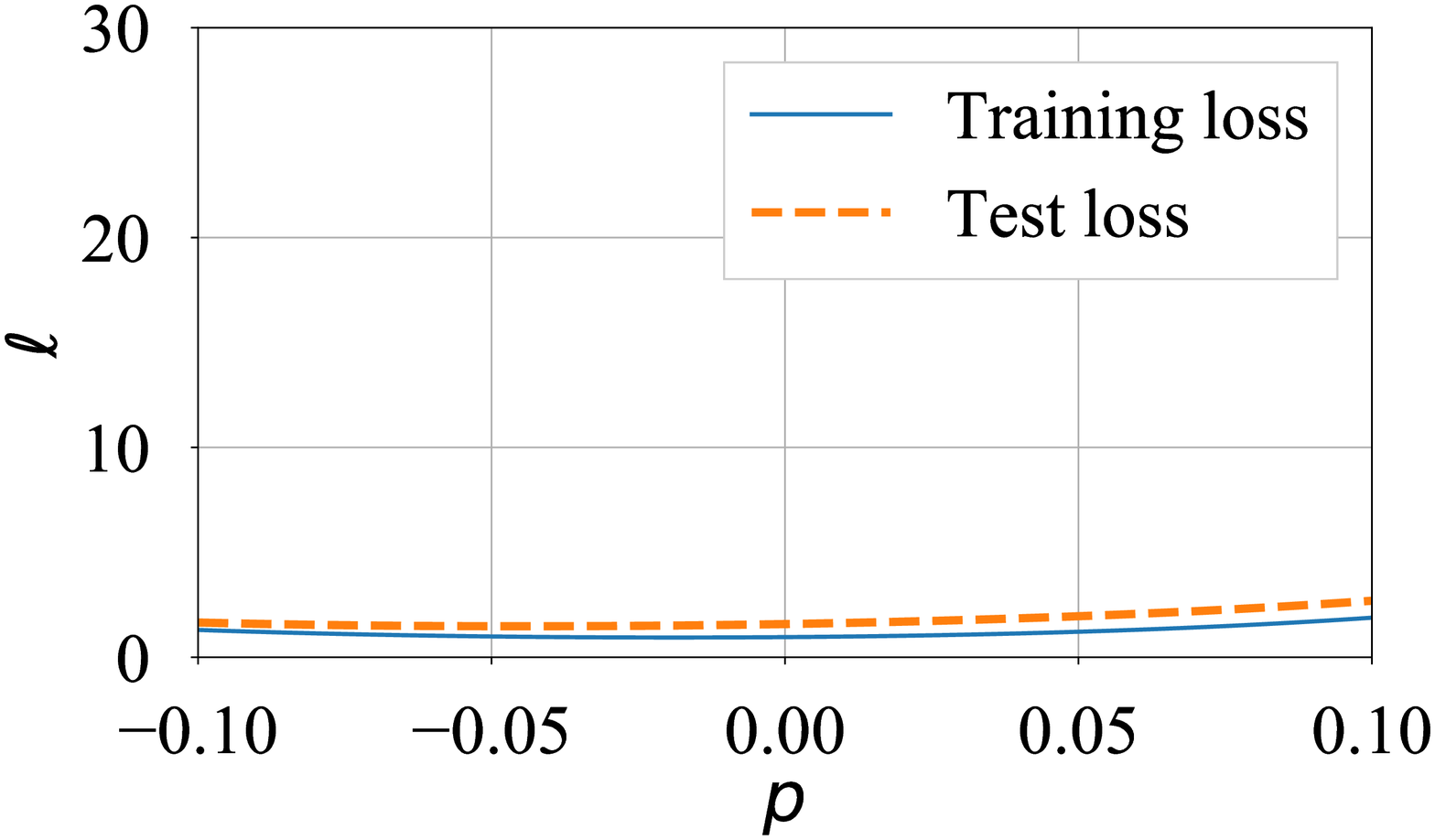}
	}
	\caption{Loss curves around the minimum obtained by VDMs for ResNet-$20$ on CIFAR-$100$ with caption of each subfigure denoting the VDM acting on the loss.}
	\label{curve_CE}
\end{figure*}

In this subsection, we proved experimental results on the flatness.
The loss surface in the deep learning is a high-dimensional hypersurface, which is difficult to draw on $2$D plane directly. Literature \cite{li2018visualizing} provides the filter normalization method for visualizing the loss landscape. Besides, \cite{li2018visualizing} also shows that the loss curves visualized using different random directions are very similar for one model. In addition, the computational cost required to visualize the $3$D loss surface is much more than that to depict the loss curve. With this in mind, the conducted experiments visualize loss curves instead of loss surfaces by using the method proposed in \cite{li2018visualizing}, and it is sufficient to illustrate the nature of the loss surface.

Figure \ref{curve_CE} gives the loss curves for different deformation settings on CIFAR-$100$ (see Table \ref{LRS2} for test accuracy results). As can be seen from Fig. \ref{curve_CE_a} and \ref{curve_CE_b}, the model deformed by $\check{\ell}=0.5\ell^2$ falls into the sharp minimum, and performs worse than the original model. However, models deformed by AP and LE obtain flat minima. In Fig. \ref{curve_CE_c} to \ref{curve_CE_h}, the larger the $a_1$ or $e_2$ (it means the loss surface is deformed to be steeper at low-loss region), the flatter the minima. These observations support that the vertical-deformation-mapping-equipped optimizer filters out some sharp minima, and then finds flat minima.

\subsection{Comparison experiments}\label{ComparExp}
\begin{table}[htbp]
	\centering
	\caption{Top-$1$ test accuracies of various learning rates and VDMs on CIFAR-$100$ for ResNet-$20$. Step decay is used for all models. The base learning rate are $1 \eta=0.1$ (also used for AP and LE), $2 \eta=0.2$, etc. Although the performance of AP and LE VDMs are different, most of them outperform the original model.}
	\begin{tabular}{llllllllll}
		\toprule
		\multirow{1}{*}Method& Top-$1$ acc. (\%) \\
%		\multirow{1}{*}      &\\
		\midrule
		 $0.5 \eta$              &$69.31$\\
		$1 \eta$               &$\textbf{69.76}$\\
		$1.5 \eta$ 			&$69.61$\\
		$2 \eta$               &$69.10$\\
		$3 \eta$               &$67.67$\\
		$4 \eta$               &$66.65$\\
		$5 \eta$               &$63.97$\\
%		$6 \eta$               &$62.95$\\
%		$10 \eta$               &$53.08$\\
		\midrule
		$\check{\ell}= 0.5\ell^2$  &$69.60$\\
		LE $(1, 0.60)$   		&$\textbf{70.22}$\\
		LE $(1, 0.70)$   		&$\textbf{69.80}$\\
		LE $(1, 0.90)$   		&$\textbf{69.92}$\\
		LE $(1, 0.99)$   		&$\textbf{70.03}$\\
		%		AP $(2, 1, 2, 1)$       &$69.53$\\
		AP $(5, 1, 2, 1)$       &$69.60$\\
		AP $(18/\pi, 1, 2, 1)$  &$\textbf{70.67}$\\
		AP $(6, 1, 2, 1)$       &$\textbf{69.82}$\\
		AP $(10, 1, 2, 1)$       &$69.19$\\
		\bottomrule
	\end{tabular}
%	\caption{Top-$1$ test accuracies of ResNet-$20$ with wide range learning rates and loss-surface-deformed ResNet-$20$ on CIFAR-$100$.}
	\label{LRS2}
\end{table}
\begin{table*}[htbp]
	\centering
	\caption{Top-$1$ test accuracies on CIFAR-$10$ and CIFAR-$100$ with VDMs.}
	\begin{tabular}{llllllllll}
		\toprule
		\multirow{2}*{} Model& \multicolumn{2}{l}{CIFAR-$10$} &\multicolumn{2}{l}{CIFAR-$100$}\\
		\cmidrule(lr){2-3}\cmidrule(lr){4-5}
		&Original  &Deformed &Original  &Deformed\\
		\midrule
		PreResNet-$20$       &$92.08$ &$\mathbf{92.64}$      [LE $(1, 0.99)$] &$67.03$ &$\mathbf{68.49}$ [AP $(2, 1, 2, 1)$]\\
		ResNet-$20$          &$92.04$ &$\mathbf{92.29}$      [LE $(1, 0.9)$] &$69.76$ &$\mathbf{70.67}$  [AP $(18/\pi, 1, 2, 1)$]\\
		PreResNet-$110$      &$94.24$ &$\mathbf{94.43}$      [LE $(1, 0.9)$] &$72.96$ &$\mathbf{73.63}$ [AP $(2, 1, 2, 1)$]\\
		ResNet-$110$          &$93.73$ &$\mathbf{94.19}$     [LE $(1, 0.8)$] &$73.93$ &$\mathbf{74.14}$ [AP $(2, 1, 2, 1)$]\\
		%DenseNet-BC-$100$    &$94.99$ &$\mathbf{95.13}$      [LE $(1, 0.99)$] &$77.38$ &$\mathbf{77.40}$ [AP $(2, 1, 2, 1)$]\\
		EfficientNet-B$0$     &$94.89$ &$\mathbf{94.91}$     [AP $(10, 1, 1, 1)$] &$77.34$ &$\mathbf{78.53}$ [AP $(18/\pi, 1, 2, 1)$]\\
		EfficientNet-B$1$    &$94.49$ &$\mathbf{94.87}$      [AP $(1.5, 0.7, 1, 0)$] &$78.33$ &$\mathbf{79.36}$ [AP $(1.5, 0.7, 1, 0)$]\\
		SE-ResNeXt-$29$ &$96.15$ &$\mathbf{96.39}$  [AP $(10, 1, 1, 1)$] &$83.65$&$\mathbf{83.83}$ [LE $(1, 0.3)$]\\
		\bottomrule
	\end{tabular}
	\label{Comparison_CF}
\end{table*}

\begin{table}[tbp]
	\centering
	\caption{Comparisons of classification performance (\%, \textbf{1-crop testing}) on ImageNet validation
		set.}
	\begin{tabular}{llllllllll}
		\toprule
		\multirow{2}*{} Model& \multicolumn{2}{c}{Top-$1$ acc. (\%)} &VDM \\
		%&Epoch&Batch size\\
		\cmidrule{2-3}
		&Original  &Deformed \\
		\midrule
		PreResNet-$18$       &$70.03$ &$\mathbf{70.46}$ &LE $(1, 1.9)$ \\
%		&120 &256\\
		ResNet-$18$       &$70.19$ &$\mathbf{70.43}$ &LE $(1, 0.5)$ \\
%		&120 &256\\
		ResNet-$34$          &$73.51$ &$\mathbf{73.61}$  &LE $(1, 0.6)$ \\
%		&120 &256\\
		%DenseNet-$121$        &$74.09$ &$\mathbf{74.11}$  &AP $(1, 0.5, 2, 0.3)$ &90 &64\\
		\bottomrule
	\end{tabular}
	\label{Comparison_IN}
\end{table}
In this subsection, we conduct comparison experiments.  
%For all of the experiments, SGDM is chosen as the optimizer with step decay scheduler without dropout. 
First, we compare various learning rates and VDMs on CIFAR-$100$ for ResNet-$20$ and report the results in Table \ref{LRS2}. The results show that AP and LE are indeed better than the learning rate multiplied by fixed values, which shows that the improvement of AP and LE is not because the learning rate of the original model is not adequate. In addition, Table \ref{LRS2} also shows that most models equipped with AP or LE are better than the original model, which shows that AP or LE does not necessarily require specific parameters (such as $a_1 $ and $e_1 $) to perform well. Note that $\check{\ell} = 0.5 \ell^2$ and AP $(10, 1, 2,1)$ are not commonly used deformation mappings (the former is used to provide an example that runs counter to AP and LE, and the parameters $a_1$ and $a_3$ of the latter are too extreme). The results of $\check{\ell} = 0.5 \ell^2$ and AP $(10, 1, 2,1)$ are only used to supplement Fig. \ref{curve_CE}.

%comparisons among learning-rate-scheduler-equipped CNNs and loss-surface-deformed CNNs, and comparisons among loss-surface-deformed CNNs and the original CNNs are performed. 

The comparison experiments among original models and deformed models are conducted on CIFAR-$10$, CIFAR-$100$, and ImageNet \cite{Krizhevsky09learningmultiple, russakovsky2015imagenet}. %Firstly, the comparisons among the state-of-the-art (SOTA) learning rate scheduler (such as hyperbolic-tangent decay which is abbreviated as HTD \cite{hsueh2019stochastic}) and VDMs are conducted. 
%The step decay scheduler is used for models equipped with VDMs, while other hyperparameters remain the same among all models. The results are shown in Table \ref{LRS}. 
%DenseNet-BC-$100$ \cite{huang2017densely}, 
Specifically, original EfficientNet-B$0$ \cite{tan2019efficientnet}, EfficientNet-B$1$ \cite{tan2019efficientnet}, PreResNet-$110$ \cite{he2016identity}, PreResNet-$20$ \cite{he2016identity}, ResNet-$110$ \cite{He2016Deep}, ResNet-$20$ \cite{He2016Deep} and SE-ResNeXt-$29$ ($16\times64$d) \cite{Hu_2018_CVPR} with their deformed versions are compared on CIFAR-$10$ and CIFAR-$100$, with the same hyperparameters.
We evaluate all the models by median test accuracies of $5$ runs on CIFAR. 
The comparisons are also conducted on ImageNet, and only the lightweight models such as ResNet-$18$, PreResNet-$18$, and ResNet-$34$ are involved, due to the limitation of computing power. ResNet-$18$, PreResNet-$18$, and ResNet-$34$ are trained for $120$ epochs on ImageNet, and the batch size is $256$.
As indicated in Table~\ref{Comparison_CF} and \ref{Comparison_IN}, all of the models with VDMs performs better than the original models. Under the same training settings, the accuracy between the original model and the deformed model of CIFAR-10 increased by up to $0.76\%$ on CIFAR-$10$, $1.46\%$ on CIFAR-$100$, and $0.43\%$ on ImageNet, respectively. These improvements are significant for CNNs, especially without introducing any additional tricks.

\section{Conclusions}\label{Conclusion}
In this paper, the novel concept of deforming the loss surface to affect the behaviour of the optimizer has been proposed with its mathematical form given. Several VDMs have been investigated for their contributions to deforming the loss surface. Theoretical analyses demonstrate that the proposed VDMs have the capability to filter out sharp minima. Moreover, relevant numerical experiments have been organized to visually illustrate the superiority of the proposed VDMs. Besides, the performance of finding flat minima for the original SGDM and VDM-enhanced SGDM have been evaluated on CIFAR-$100$ with the actual loss curve visualization. On CIFAR-$10$ and CIFAR-$100$, popular deep learning models enhanced by the proposed method in our experiments have shown superior test performance than the original model with up to $1.46\%$ improvement. Overall, the theoretical analyses, simulations, and the related comparative experiments have illustrated the effectiveness and potential of the proposed deformation method in improving the generalization performance. In addition to VDM which aims at filtering sharp minima, more forms of deformation mappings will be studied in the future to address other problems in deep learning, such as the saddle point problem and gradient explosion problem.
%\section{Broader impact}\label{impact}
%In recent years, the SOTA models in deep learning often involve high computing resource overheads, which means that training these models may bring a lot of energy consumption or carbon emissions. Unlike structure-based approaches, using VDMs to improve performance does not introduce significant additional computational overhead, which means the same result can be obtained by lower power consumption. Besides, deforming the loss surface is a novel perspective for better performance. The method of deformation is not only the deformation-function-based method, and the effect of deformation is not only to promote the optimizer to find a flat solution. However, like most methods of promoting computers to replace part of human labour, this work may increase unemployment slightly, until finding a living art that can cope with the automation course.
%\section*{References}
%\bibliographystyle{IEEEtran}
\bibliographystyle{aaai}
%\bibliographystyle{unsrt}
%\normalem
\bibliography{Reference}
\newpage
\setcounter{page}{1}
\part*{\centering Deforming the Loss Surface to Affect the Behaviour of the Optimizer\\
	(Supplementary Material) }
%\title{Deforming the Loss Surface\\
%(Supplementary Material) }

	In the supplementary material, the illustration of the deformation method is provided, proofs of Theorems are given, and details of simulations and experiments are presented. Besides, additional experiments are conducted.
\appendix
\section{Sketch of the deformation method}

\begin{figure}[htbp]
	\centering
	
	\subfigure[]{
		\label{Sketch2}
		\includegraphics[width=2.6in]{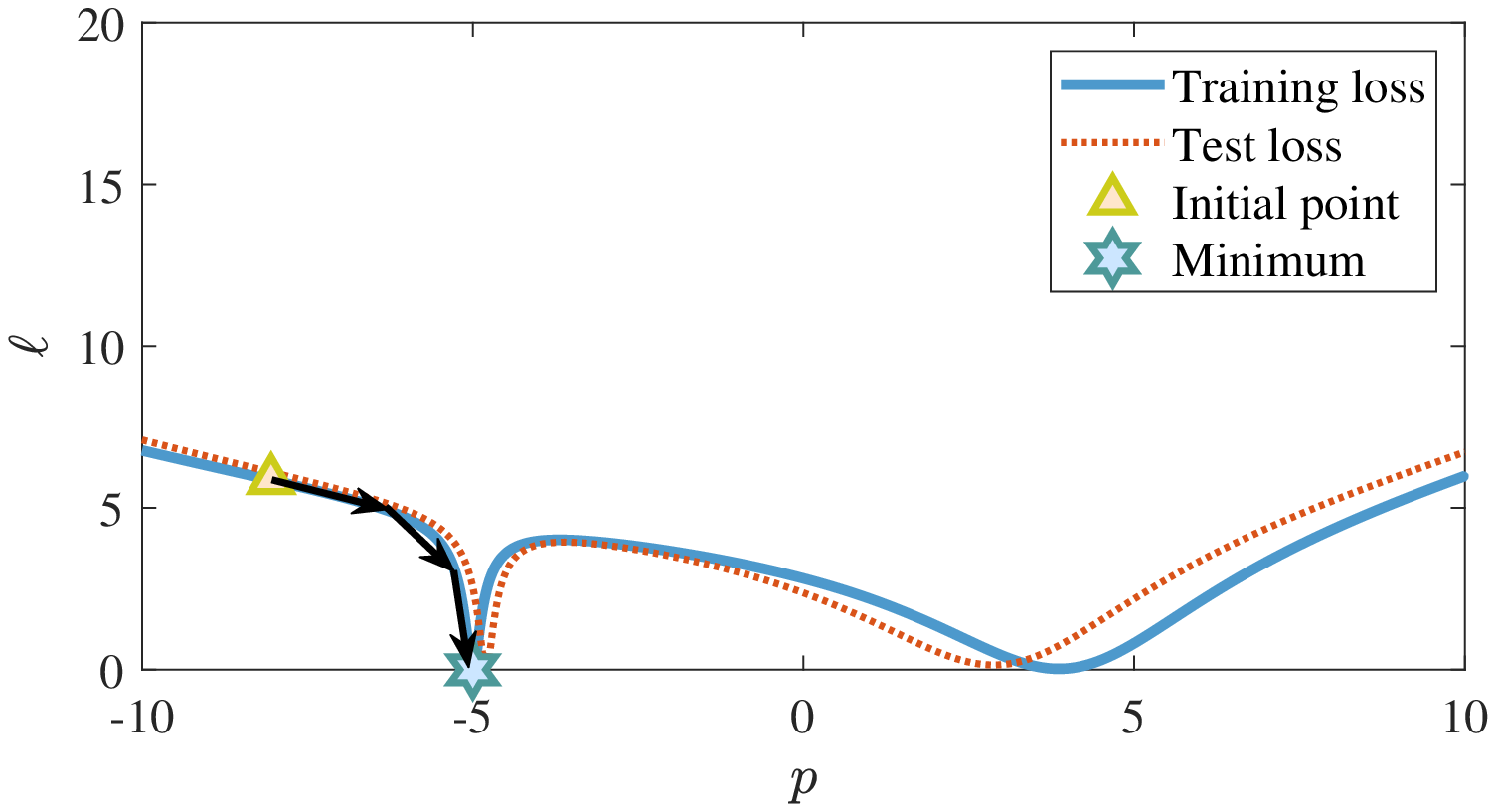}
	}
	\subfigure[]{
		\label{Sketch3}
		\includegraphics[width=2.6in]{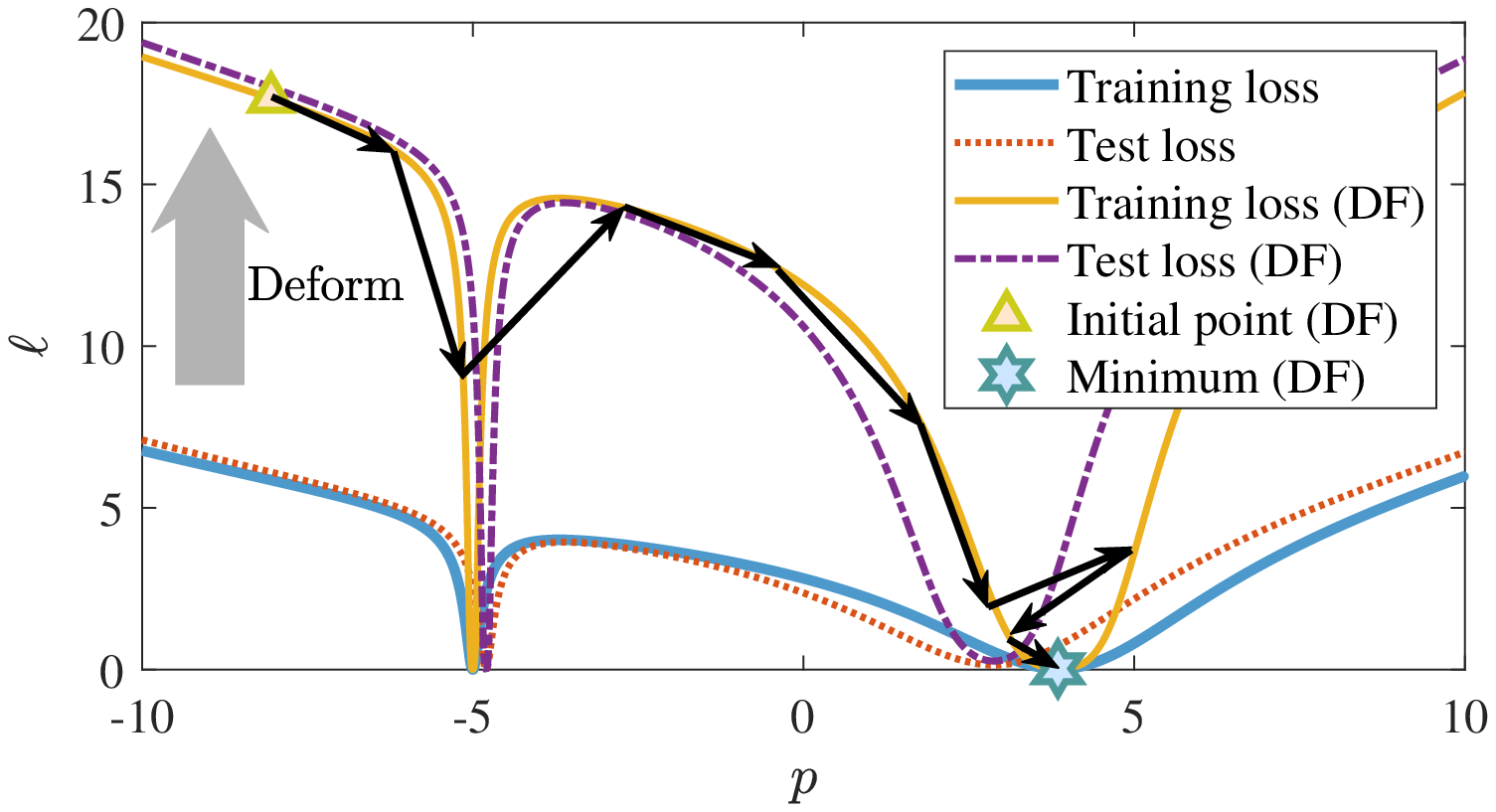}
	}
	\caption{Sketches of the flat minimum filter. (a) The original loss curve. (b) The deformed loss curve.}
	%        \caption{Sketches of the flat minimum filter. (a) If the parameter updating procedure starts with the initial parameter on the original loss curve, the sharp minimum is obtained finally. (b) Taking the same initial parameter, the sharp minimum is skipped, and the flat minimum is obtained instead.}
	%%%    \caption{Sketches of the flat minimum filter. In the subfigures, the thin blue solid line denotes the original training loss, and thin orange dashed line denotes the original test loss, the thick yellow solid line denotes the deformed training loss, thick purple dashed line denotes the deformed test loss, $p_1$ is parameter gives the sharp minimum in training, $p_2$ is parameter give the flat minimum in testing, and $(p_5, \ell^*(p_5))$ S is the extreme point where the second derivative is less than zero. (a) shows that the sharp minimum leads to higher test loss. If the parameter updating procedure starts at $p_3$ in the original loss curve, the sharp minimum is obtained finally, as shown in (b). However, if the loss curve is deformed as (c) and (d) shows, take $(p_3, \ell^*(p_3))$ as the initial point, the sharp minimum is ignored. Whether $(p_3, \ell^*(p_3))$ or $(p_6, \ell^*(p_6))$ is taken as the initial points, the flat minimum is obtained for the deformed loss curve. }
	\label{Sketch}
\end{figure}

Sketches of the flat minimum filter are shown in Fig. \ref{Sketch}. In this figure, the solid blue line denotes the original training loss; the orange dashed line denotes the original test loss; the solid yellow line denotes the deformed training loss; the purple dashed line denotes the deformed test loss. From Fig. \ref{Sketch2}, it can be seen that the sharp minimum leads to higher test loss than the flat one, and the sharp minimum is obtained finally for the original loss curve. However, if the loss curve is deformed as Fig. \ref{Sketch3} shows, the sharp minimum is escaped with the same initial parameter.
\section{Proofs}
In the following subsections, Theorem 1 and Theorem 2 are proved.
\subsection{Proof of Theorem 1}
\begin{proof}
	The ($r,s$)th element of $\check{\mathbf{H}}$ can be written as
	\[
	\mathbf{H}_{rs}^* = \frac{\partial^2 \delta(\ell)}{\partial p_r \partial p_s} = \frac{\partial \delta}{\partial \ell} \frac{\partial^2 \ell}{\partial p_r \partial p_s}=\frac{\partial \delta}{\partial \ell} \mathbf{H}_{rs}.
	\]
	Thus,
	\[
	\check{\mathbf{H}} = \frac{\partial \delta}{\partial \ell} \mathbf{H}.
	\]
	Considering the definition of eigenvalues, the following equation is obtained:
	\[
	\lambda(\check{\mathbf{H}}) = \frac{\partial \delta}{\partial \ell} \lambda(\mathbf{H}).
	\]
	The proof is thus completed.
\end{proof}
\subsection{Proof of Theorem 2}
\begin{proof}
	The parameter $p^{(k+1)}$ in step $k$ can be obtained by the following iterative rule:
	\begin{equation}\label{sgd1}
	\Delta p^{(k+1)} := - \eta \nabla \delta((\ell(p))^{(k)}),
	\end{equation}
	and
	\begin{equation}\label{sgd2}
	p^{(k+1)} := p^{(k)} + \Delta p^{(k+1)},
	\end{equation}
	where $\eta$ is the learning rate.
	%Consider dimension $n$ only, then the step size is written as $m \Delta p^{(k)} - \eta \frac{\partial \delta((\ell(p))^{(k)})} {\partial p^{(k)}}$.
	%\begin{definition}
	%Assume that within an interval $(x^{[a]}, x^{[b]})$, $f(x) : \mathbb{R} \rightarrow \mathbb{R}$ is differentiable, there are only three extreme point
	%$(x^\mathrm{[min]}$, $f^\mathrm{[min]})$ and $(x^\mathrm{[max_1]}, f^\mathrm{[max_1]})$ with $\partial^2 f^\mathrm{[min]} / \partial (x^\mathrm{[min]})^2>0$, $\partial f^\mathrm{[max_1]} / \partial (x^\mathrm{[max_1]})^2<0$, $\partial f^\mathrm{[max_2]} / \partial (x^\mathrm{[max_2]})^2<0$, $x^\mathrm{[max_1]}<x^\mathrm{[max_2]}$, then $(x^{[a]}, x^{[b]})$ is called local concave region (or ``local M region'' in the sight of geometrical shape) for $f(x)$.
	%\end{definition}
	Assume that there are only three extreme point
	$(p^\mathrm{[min]}$, $\ell^\mathrm{[min]}_n)$, $(p^\mathrm{[max_1]}, \ell^\mathrm{[max_1]}_n)$, and $(p^\mathrm{[max_2]}, \ell^\mathrm{[max_2]}_n)$, with $\partial^2 \ell^\mathrm{[min]} / \partial (p^\mathrm{[min]})^2>0$, $\partial \ell^\mathrm{[max_1]} / \partial (p^\mathrm{[max_1]})^2<0$, $\partial \ell^\mathrm{[max_2]} / \partial (p^\mathrm{[max_2]})^2<0$, $p^\mathrm{[max_1]}<p^\mathrm{[max_2]}$, and the total number of the steps is sufficiently large, the following results are achieved:
	\begin{itemize}
		\item  For the case of $p^{(k)} \in [p^\mathrm{[max_1]}, p^\mathrm{[min]}]$, if $\exists p^{(k)}$ such that
		\begin{equation}\label{less}
		\frac{\partial (\ell(p))^{(k)}} {\partial p^{(k)}} < \frac{p^{(k)} - p^\mathrm{[max_2]}  }{\eta \frac{\partial \delta}{\partial \ell} },
		\end{equation}
		and then $p^{(k+1)} > p^\mathrm{[max_2]}$, which means that the parameter will escape from $(p^{[a]}_n, p^{[b]}_n)$ if $\tilde{k}$ is sufficiently large. One can derive that if $p^{(k+1)} \in [p^\mathrm{[max_1]}, p^\mathrm{[max_2]}]$, and then $\forall p^{(k)}$,
		\begin{equation}\label{lower}
		0 \geq \frac{\partial (\ell(p))^{(k)}} {\partial p^{(k)}} \geq \frac{p^{(k)} - p^\mathrm{[max_2]} }{\eta \frac{\partial \delta}{\partial \ell} } \geq \frac{p^\mathrm{[max_1]} - p^\mathrm{[max_2]} }{\eta \frac{\partial \delta}{\partial \ell} }.
		\end{equation}
		\item  For the case of $p^{(k)} \in [p^\mathrm{[min]}, p^\mathrm{[max_2]}]$, if $\exists p^{(k)}$ such that
		\begin{equation}\label{more}
		\frac{\partial \delta((\ell(p))^{(k)})} {\partial p^{(k)}} > \frac{p^{(k)} - p^\mathrm{[max_1]} }{\eta \frac{\partial \delta}{\partial \ell}},
		\end{equation}
		and then $p^{(k+1)} < p^\mathrm{[max_1]}$, which means that the parameter will escape from $(p^{[a]}_n, p^{[b]}_n)$ if $\tilde{k}$ is sufficiently large. One can derive that if $p^{(k+1)} \in [p^\mathrm{[max_1]}, p^\mathrm{[max_2]}]$, then $\forall p^{(k)}$,
		\begin{equation}\label{upper}
		0 \leq \frac{\partial (\ell(p))^{(k)}} {\partial p^{(k)}} \leq \frac{p^{(k)} - p^\mathrm{[max_1]} }{\eta \frac{\partial \delta}{\partial \ell} } \leq \frac{p^\mathrm{[max_2]} - p^\mathrm{[max_1]} }{\eta \frac{\partial \delta}{\partial \ell} }.
		\end{equation}
		%  \item  For the case of $p^{\mathrm{[max]}}_n < p^{\mathrm{[min]}}_n$, if $\exists p^{(k)} \in [p^\mathrm{[max]}_n, p^{[b]})$, $m \Delta p^{(k)} - \eta \frac{\partial \delta((\ell(p))^{(k)})} {\partial p^{(k)}} < p^\mathrm{[max]}_n - p^{(k)}$, then $p^{(k+1)} < p^\mathrm{[max]}_n$.
		%  \item  For the case of $p^{\mathrm{[max]}}_n > p^{\mathrm{[min]}}_n$, if $\exists p^{(k)} \in (p^{[a]}, p^\mathrm{[max]}_n]$, $m \Delta p^{(k)} - \eta \frac{\partial \delta((\ell(p))^{(k)})} {\partial p^{(k)}} > p^\mathrm{[max]}_n - p^{(k)}$, then $p^{(k+1)} > p^\mathrm{[max]}_n$.
	\end{itemize}
	Let $\triangle p^\mathrm{[max]} = p^\mathrm{[max_2]} - p^\mathrm{[max_1]}$ represents the projection of the distance between adjacent maximum values on $p$.
	Based on this consideration, $\triangle p^\mathrm{[max]}$ can be regarded as an independent variable that does not depend on $p$ or ${\partial (\ell(p))^{(k)}}/{\partial p^{(k)}}$. For a sufficiently large $\tilde{k}$, if $p^{(k+1)}$ remains in $(p^{[a]}, p^{[b]})$, together with inequalities \eqref{lower} and \eqref{upper}, $\partial (\ell(p))^{(k)} / \partial p^{(k)}$ is bounded by
	\begin{equation}\label{bounded}
	\bigg| \frac{\partial (\ell(p))^{(k)}} {\partial p^{(k)}}\bigg| \leq
	\frac{\triangle p^\mathrm{[max]}}{\eta \frac{\partial \delta}{\partial \ell} }.
	\end{equation}
	The proof is thus completed.
\end{proof}
%%%Thus, $(\ell(\boldsymbol{p}))^{(k)}$ meets Lipschitz condition:
%%%\begin{equation}\label{lp}
%%%||(\ell(\boldsymbol{p}))^{(k)}-\ell(\hat{\boldsymbol{p}}^{(k)})|| \leq \frac{1}{\eta \frac{\partial \delta}{\partial \ell} }\left(\sum_{n=1}^{\tilde{n}} \left({j_n}\right)^{-2}\right)^{-\frac{1}{2}} { ||\boldsymbol{p}^{(k)} - \hat{\boldsymbol{p}}^{(k)}||},
%%%\end{equation}

%\appendix
\section{Details of simulations and additional simulations}
\begin{figure}[htbp]
	\centering
	\subfigure[Original (filled contour view)]{
		\label{complex_sim_a}
		\includegraphics[width=2.4in]{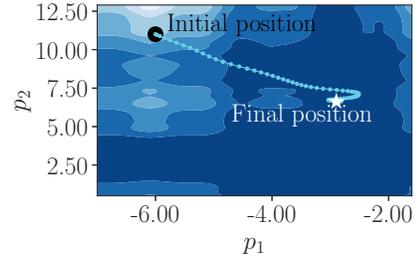}
	}
	\subfigure[Deformed (filled contour view)]{
		\label{complex_sim_b}
		\includegraphics[width=2.4in]{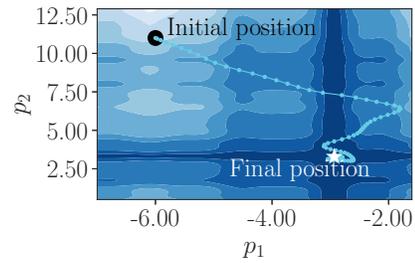}
	}
	\quad    %? \quad ???
	\subfigure[Original (3D view)]{
		\label{complex_sim_c}
		\includegraphics[width=2.5in]{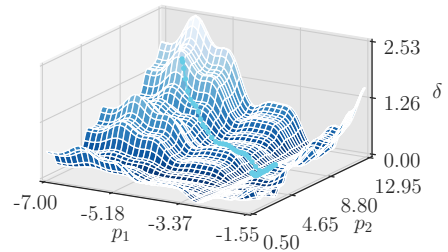}
	}
	\subfigure[Deformed (3D view)]{
		\label{complex_sim_d}
		\includegraphics[width=2.5in]{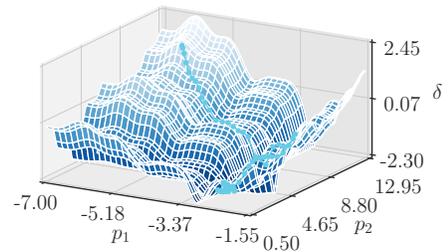}
	}
	%    \quad    %? \quad ???
	%    \subfigure[]{
	%    \label{complex_sim_e}
	%    	\includegraphics[width=2.5in]{ori_momentum_complex_2_2Db.eps}
	%    }
	%    \subfigure[]{
	%    \label{complex_sim_f}
	%	\includegraphics[width=2.5in]{logexp90_momentum_complex_2_2Db.eps}
	%    }
	%\quad    %? \quad ???
	%    \subfigure[]{
	%    \label{complex_sim_g}
	%    	\includegraphics[width=2.5in]{ori_momentum_complex_2_3D.eps}
	%    }
	%    \subfigure[]{
	%    \label{complex_sim_h}
	%	\includegraphics[width=2.5in]{logexp90_momentum_complex_2_3D.eps}
	%    }
	\caption{Deformation on a simulated loss surface in a complex case with two parameters.}
	\label{complex_sim}
\end{figure}
First,  details of simulation in Fig. \ref{filter_sim} is given: the function of the loss surface is $\ell(p_1, p_2) = (0.2 (0.01 (p_1+0.5)^2+1.32 \arctan(0.5 p_1-2)^2+\arctan(10 (p_1+5))^2)-0.5232)
((0.02 (p_2+0.5)^2 +2.64 \arctan(0.5 p_2-2)^2 + 2 \arctan(10 (p_2+5))^2) -5.232)$; the VDM is AP$(10, 1, 1, 1)$; the initial parameters are $p_1 = -9.5$ and $p_2 = -10$; GDM is taken as the optimizer and is trained for $400$ epochs; the learning rate is $0.02$, and the momentum is $0.9$.
Then, we describe settings for the simulation presented by  Fig. \ref{eig_round}.  The original loss surface is determined by $\ell(p_1, p_2) =0.132 (\ln(\sin(4.7 \sin(x))+1.2)+\ln(\sin(4.7 \sin(p_2))+1.2))+0.01 (\sin(10 p_1) \sin(6 p_2))+0.001 p_1^2+0.001p_2^2 + 0.426931992284953$; the VDM is LE $(1, 0.99)$; the initial parameters are randomly initialized in  $[-10 , 10]$ for both $p_1$ and  $p_2$; GD is taken as the optimizer and is trained for $300$ epochs with step decay (learning rate is initialized as $0.2$, and is multiplied by $0.1$ at epoch $150$ and $225$).

Moreover, simulations on complex situation is also performed and presented in Fig. \ref{complex_sim}. The initial parameters are taken as $p_1 = -6$, and $p_2 = 11$. The minimum obtained by the original optimizer is sharp in one direction with final loss value $0.005$, as shown in Fig. \ref{complex_sim_a} and \ref{complex_sim_c}. However, after deforming the loss surface, the minimum obtained by the optimizer is flat in all directions with the loss value $1.19\times10^{-5}$, as shown in Fig. \ref{complex_sim_b} and \ref{complex_sim_d}. The function of the loss surface is taken as $\ell(p_1, p_2) = 0.12((0.1p_1)^2+\cos(p_1)+\sin(3p_1)/3+\cos(5p_1)/5+\sin(7p_1)/7+1.331)
((0.158p_2)^2+\cos(p_2)+\sin(3p_2)/3+\cos(5p_2)/5+\sin(7p_2)/7+1.25)$, and the VDM is $\delta(\ell) = \ln(\exp(\ell-0.9))$. The GDM is taken as the optimizer and trained for $100$ epochs with the learning rate $0.02$ and the momentum $0.9$. The results from Fig. \ref{complex_sim} demonstrate the superiorities of the proposed VDM: the vertical-deformation-mapping-equipped optimizer escapes from the sharp minimum and finds the flat region.

\section{Details of  Experiments}

\begin{table*}[htbp]
	\centering
	\caption{Settings of the experiments on CIFAR-$10$ and CIFAR-$100$.}
	\begin{tabular}{llllllllll}
		\toprule
		\multirow{2}*{} Model& Epoch &Batch  &Initial &Milestones  &Weight &Distributed\\
		&  & size & learning rate & &decay &training\\
		% \cmidrule{2-3}
		\midrule
		PreResNet-$20$       &250 &128 &0.1&\{100, 150, 200\}& 0.0001 &No\\
		ResNet-$20$          &250 &128 &0.1&\{150, 225\}& 0.0005 &No\\
		PreResNet-$110$      &250 &128 &0.1&\{100, 150, 200\}& 0.0001 &No\\
		ResNet-$110$          &250 &128 &0.1&\{150, 225\}& 0.0005 &No\\
		%DenseNet-BC-$100$    &300 &64 &0.1&\{150, 225\}& 0.0001 &No\\
		EfficientNet-B$0$     &250 &128 &0.1&\{150, 225\} & 0.0001 &No\\
		EfficientNet-B$1$    &250 &128 &0.1&\{150, 225\} & 0.0001 &No\\
		SE-ResNeXt-$29$ &300 &128 &0.1&\{150, 225\} &  0.0005 &No\\
		\bottomrule
	\end{tabular}
	\label{C10_setting}
\end{table*}
\begin{table*}[htbp]
	\centering
	\caption{Settings of the experiments on ImageNet.}
	\begin{tabular}{llllllllll}
		\toprule
		\multirow{2}*{} Model& Epoch &Batch  &Initial &Milestones&Weight&Distributed  &Number \\
		&  & size & learning rate &&decay& training & of GPUs \\
		% \cmidrule{2-3}
		\midrule
		ResNet-$18$       &120 &256 &0.2&\{30, 60, 90\}&0.0001&Yes&2\\
		PreResNet-$18$       &120 &256 &0.2&\{30, 60, 90\}&0.0001&Yes&2\\
		ResNet-$34$          &120 &256 &0.2&\{30, 60, 90\}&0.0001&Yes&2\\
		%DenseNet-$121$          &90 &64 &0.1&\{30, 60\}&0.0001&Yes&2\\
		
		\bottomrule
	\end{tabular}
	\label{IN_setting}
\end{table*}
In this section, settings of experiments are provided, including descriptions of datasets, hyperparameters, and computation hardware information.

Both CIFAR-$10$ and CIFAR-$100$ contain $60K$ images, where $50K$ for training and $10K$ for testing. The splitting of the training set and the test set follows the common settings of the Python version of CIFAR \cite{Krizhevsky09learningmultiple}.
%% (downloading link: \href{http://www.cs.toronto.edu/~kriz/cifar.html}{http://www.cs.toronto.edu/~kriz/cifar.html}). 
The ImageNet dataset used in this paper contains $1000$ classes, with $1.28M$ training images and $50K$ validation images.
%% (downloading link: \href{http://image-net.org/download}{http://image-net.org/download}). 
The splitting of the training and test set for ImageNet follows the common settings \cite{shen2019meal, huang2017snapshot, zhang2017polynet}. The ImageNet images are randomly cropped into $224 \times 224$. In the validation set of ImageNet, the single $224 \times 224$ center crop is carried out for testing.  
Table \ref{C10_setting} and \ref{IN_setting} show the experimental settings of ImageNet and CIFAR, respectively. In these tables, the set of some specific epochs are denoted as milestones, and the learning rate is multiplied by $0.1$ when these epochs are reached. 
%For PreResNets, Se-ResNext-$29$-$16\times64$d, DenseNet-BC-$100$ on CIFAR, the original code is from \cite{bigballon2019cifarzoo}; for EfficientNets, the original code is from \cite{pim2020}; for ResNets on CIFAR, the code is from \cite{VLLNN}; for ImageNet models, the code is from \cite{PIC}.
The number of GPUs in distributed training may affect accuracy. Thus the number of GPUs are also listed in Table \ref{IN_setting}. Our computing infrastructure contains two servers, one with $10$ GeForce RTX $2080$ Ti GPUs (with $11$GB RAM per GPU), and the other with $2$ Quadro RTX $8000$ GPUs (with $48$GB RAM per GPU). However, since not all computing resources can be accessed at all times, some experiments are conducted on $2$ or $4$ GPUs. The experiments are performed by using PyTorch $1.4.0$ and CUDA $10.0$ within Python $3.7.6$.

%\section{Additional experiments}
%In this section, additional experiments are conducted. Specifically, comparison of top-$1$ test accuracies ResNet-$20$ with wide range learning rates and loss-surface-deformed ResNet-$20$ on CIFAR-$100$ are described in Table \ref{LRS2}. Although the learning rates are taken from a wide range, ResNet-$20$ based on AP $(18/\pi, 1, 2, 1)$, LE $(1, 0.6)$, and LE $(1, 0.9)$ performs better.

\section{Training time consumption}

The runtime differences among each deformed models and the original models are insignificant. For example, on ImageNet, the average training time costs of each epoch for the original ResNet-$34$ and the ResNet-$34$ equipped with LE $(1, 0.6)$ VDM are $1202.71$ s and $1192.77$ s, respectively. The average validation time costs of each epoch for the original ResNet-$34$ and the ResNet-$34$ equipped with LE $(1, 0.6)$ VDM are $36.22$ s and $36.29$ s, respectively.

\end{document}